\documentclass[12pt]{report}
 
\title{Towards High-Frequency Tracking and Fast Edge-Aware Optimization}

\author{Akash Bapat}
\date{}

\usepackage[nopostdot,nogroupskip, style=super, nonumberlist]{glossaries}
\renewcommand{\glossarysection}[2][]{}
\makeglossaries
\newacronym{dts}{DTS}{domain transform solver}
\newacronym{dt}{DT}{domain transform}
\newacronym{arvr}{AR/VR}{augmented reality/virtual reality}
\newacronym{imu}{IMU}{inertial measurement unit}
\newacronym{rs}{RS}{rolling shutter}
\newacronym{ccd}{CCD}{charge-coupled device}
\newacronym{cmos}{CMOS}{complementary metal oxide semiconductor}
\newacronym{gs}{GS}{global shutter}
\newacronym{irls}{IRLS}{iteratively reweighted least squares}
\newacronym{svd}{SVD}{singular value decomposition}
\newacronym{ar}{AR}{augmented reality}
\newacronym{vr}{VR}{virtual reality}
\newacronym{dslr}{DSLR}{digital single-lens reflex}
\newacronym{fps}{fps}{frames-per-second}
\newacronym{ode}{ODE}{ordinary differential equation}
\newacronym{dof}{DoF}{degree of freedom}
\newacronym{lidar}{LiDAR}{light detection and ranging}
\newacronym{laser}{LASER}{light amplification by stimulated emission of radiation}
\newacronym{hmd}{HMD}{head-mounted display}
\newacronym{ir}{IR}{infra-red}
\newacronym{fpga}{FPGA}{field-programmable gate array}
\newacronym{ram}{RAM}{random-access memory}
\newacronym{fov}{FoV}{field of view}
\newacronym{snr}{SNR}{signal-to-noise ratio}
\newacronym{psnr}{PSNR}{peak signal-to-noise ratio}
\newacronym{pkr}{PKR}{peak ratio}
\newacronym{mpl}{MPL}{motion-to-photon latency}
\newacronym{jnd}{JND}{just noticeable difference}
\newacronym{slam}{SLAM}{simultaneous localization and mapping}
\newacronym{led}{LED}{light-emitting diode}
\newacronym{scaat}{SCAAT}{single-constraint-at-a-time}
\newacronym{cnn}{CNN}{convolutional neural net}
\newacronym{fbs}{FBS}{fast bilateral solver}
\newacronym{hfbs}{HFBS}{hardware-friendly bilateral solver}
\newacronym{asic}{ASIC}{application-specific integrated circuit}
\newacronym{pnp}{PnP}{perspective-n-point}
\newacronym{p3p}{P3P}{perspective-3-point}
\newacronym{mvs}{MVS}{multi-view stereo}
\newacronym{ba}{BA}{bundle adjustment}
\newacronym{gps}{GPS}{global positioning system}
\newacronym{sfm}{SfM}{structure-from-motion}
\newacronym{ransac}{RANSAC}{random sampling and consensus}
\newacronym{sam}{SaM}{structure-and-motion}
\newacronym{rgbd}{RGB-D}{red-green-blue-depth}
\newacronym{svm}{SVM}{support vector machine}
\newacronym{crf}{CRF}{conditional random field}
\newacronym{gpu}{GPU}{graphics processing unit}
\newacronym{bls}{BLStereo}{Bilateral Stereo}
\newacronym{slic}{SLIC}{simple linear iterative clustering}
\newacronym{mc_cnn}{MC-CNN}{matching cost CNN}
\newacronym{ycbcr}{YCbCr}{luma, chroma blue-difference, chroma red-difference}
\newacronym{mae}{MAE}{mean absolute error}
\newacronym{rmse}{RMSE}{root mean square error}
\newacronym{rgb}{RGB}{red-green-blue}
\newacronym{ssim}{SSIM}{structural similarity}
\newacronym{gt}{GT}{ground-truth}
\newacronym{mad}{MAD}{mean absolute difference}
\newacronym{sg_wls}{SG-WLS}{semi-global weighted least squares}
\newacronym{bgu}{BGU}{bilateral guided upsampling}
\newacronym{adas}{ADAS}{advanced driver assist systems}
\newacronym{ekf}{EKF}{extended Kalman filter}
\newacronym{fast}{FAST}{features from accelerated segment test}
\newacronym{lm}{LM}{Levenberg–Marquardt}
\newacronym{roi}{RoI}{region of interest}
\newacronym{ptam}{PTAM}{parallel tracking and mapping}
\newacronym{klt}{KLT}{Kanade-Lucas-Tomasi}
\newacronym{sift}{SIFT}{scale invariant feature transform}
\newacronym{surf}{SURF}{speeded-up robust features}
\newacronym{orb}{ORB}{oriented FAST and rotated BRIEF}
\newacronym{brisk}{BRISK}{binary robust invariant scalable keypoint}
\newacronym{hd}{HD}{high-definition}
\newacronym{hec}{HEC}{hand-eye calibration}
\newacronym{slerp}{SLERP}{spherical linear interpolation}
\newacronym{soc}{SoC}{system on a chip}

\usepackage[hidelinks]{hyperref}
\usepackage[numbers]{natbib}
\usepackage[inline]{enumitem}

\usepackage{xspace}
\usepackage{graphicx}
\usepackage{subcaption}
\usepackage{xcolor}
\usepackage{stackengine}
\usepackage{gensymb}

\newcommand\acrfullit[1]{\textit{\acrlong{#1}} (\acrshort{#1})}
\newcommand\whitespace{\ }

\newcommand{\eg}{{\it e.g.}\xspace}
\newcommand{\ie}{{\it i.e.}\xspace}
\newcommand{\vs}{{vs\mbox{.}}\xspace}

\newcommand\hf{high-frequency\xspace}
\newcommand\rs{\acrshort{rs}\xspace}

\newcommand{\myparagraph}{\paragraph}

\begin{document}
\maketitle

\newpage
\begin{center}
\textit{For my parents}    
\end{center}

\begin{abstract}

Computer vision has seen tremendous success in refashioning cameras from mere recording equipment to devices which can measure, understand, and sense the surroundings.
Efficient algorithms in computer vision have now become essential for processing the vast amounts of image data generated by a multitude of devices as well as enabling real-time applications like \acrfull{arvr}.
This dissertation advances the state of the art for \acrshort{arvr} tracking systems by increasing the tracking frequency by orders of magnitude and proposes an efficient algorithm for the problem of edge-aware optimization.

\acrshort{arvr} is a natural way of interacting with computers, where the physical and digital worlds coexist.
We are on the cusp of a radical change in how humans perform and interact with computing.
This has been led by major technological advancements in hardware and in the tracking, rendering, and display algorithms necessary to enable \acrshort{arvr}.
Humans are sensitive to small misalignments between the real and the virtual world, and tracking at kilo-Hertz frequencies becomes essential.
Current vision-based systems fall short, as their tracking frequency is implicitly limited by the frame-rate of the camera.
This thesis presents a prototype system which can track at orders of magnitude higher than the state-of-the-art methods using multiple commodity cameras.
The proposed system exploits characteristics of the camera traditionally considered as flaws, namely rolling shutter and radial distortion.
The experimental evaluation shows the effectiveness of the method for various degrees of motion.

Furthermore, edge-aware optimization is an indispensable tool in the computer vision arsenal for accurate filtering of depth-data and image-based rendering, which is increasingly being used for content creation and geometry processing for \acrshort{arvr}.
As applications increasingly demand higher resolution and speed, there exists a need to develop methods that scale accordingly.
This dissertation proposes such an edge-aware optimization framework which is efficient, accurate, and algorithmically scales well, all of which are much desirable traits not found jointly in the state of the art.
The experiments show the effectiveness of the framework in a multitude of computer vision tasks such as computational photography and stereo.
\end{abstract}
\newpage

\tableofcontents
\clearpage

\newpage
\phantomsection
\addcontentsline{toc}{section}{\listfigurename}
\listoffigures

\newpage
\section*{List of Abbreviations}
\addcontentsline{toc}{section}{List of Abbreviations}
\printglossary[title={}]
\clearpage

%
%
%
\chapter{Introduction}
\label{chap:introduction}
Human beings continuously measure and sense the surroundings to survive in this complex world.
Our visual system performs a crucial function by letting us `see' by capturing the light incident on the retina.
The visual system processes the captured images to provide us with depth cues, motion estimates, relative locations of different objects, recognizes different types of objects and much more, just within fractions of a second.
The field of computer vision is set out to provide all of the aforementioned capabilities like sensing, measuring and scene understanding to computers using cameras.

Traditionally, cameras were used for recording the visual aspects of the surroundings like in photographs and videos.
Computer vision has transformed cameras from just recording devices to devices which provide sensing and scene understanding capabilities.
Computer vision has created new modes of capture which combines computation and imaging in the form of computational photography, developed novel measurement/sensing methods like \acrfull{slam} to reconstruct 3D geometry and estimate ego-motion and enabled understanding of higher-level semantics like vegetation.
This thesis focuses on two aspects of computer vision algorithms: efficient and high-frequency processing of image data.
These two themes are applied to the problems of \hf tracking for \acrfull{arvr} and edge-aware optimization.

\begin{figure}
\centering
\includegraphics[width=\textwidth]{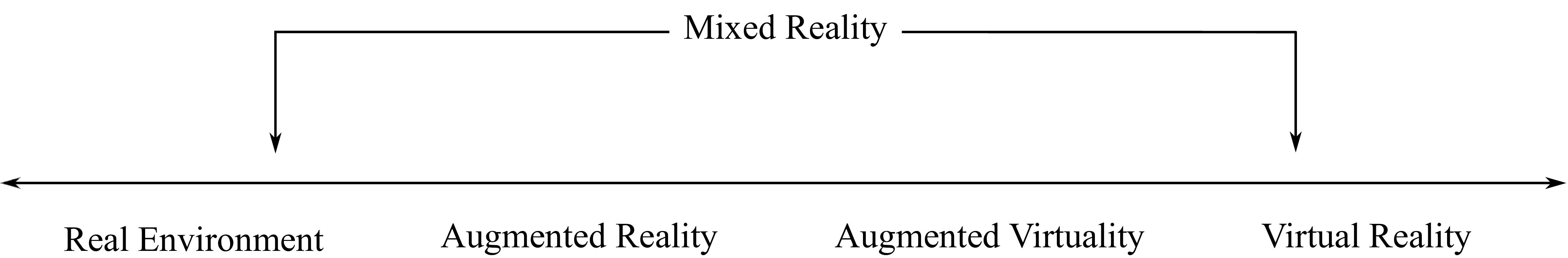}
\caption{\cite{milgram1995augmented}'s taxonomy for \acrshort{arvr} over a reality–virtuality continuum. \label{fig:arvr_continuum}}
\end{figure}

The fields of \acrshort{arvr} tackle the hard problem of fooling the human senses.
While \acrfull{vr} presents the user an entirely virtual space in which the user can interact, navigate and feel `immersed', \acrfull{ar} adds virtual content to the real world.
\cite{milgram1995augmented} provided a taxonomy called `reality–virtuality continuum' in their seminal work to organize \acrshort{ar} and \acrshort{vr} as shown in Fig. \ref{fig:arvr_continuum}.
In recent years, we have made great strides across this continuum with increasing interest by academics and the industry alike.
This resurgence is in part fueled by \acrshort{arvr}'s great promise in a wide variety of fields such as medical treatment and surgery \citep{khor2016augmented}, training \citep{boud1999virtual}, manufacturing \citep{ong2013virtual} and entertainment \citep{von2017augmented}, as well as technological advances and improved computing capabilities.

Any \acrshort{arvr} system continuously tracks the user's motion, renders the virtual content/objects according to the user's viewpoint, and then finally displays the content on a screen, so that the user believes the virtual content is located in their surrounding physical space.
To maintain a reliable registration of the virtual world with the real world, \acrshort{arvr} applications impose stringent requirements on the accuracy and speed of its core components: tracking, rendering and display \citep{lincoln2016motion}.
A convincing slight-of-hand for \acrshort{arvr} requires that the core components be performed at a faster rate than user perception \citep{sanchez2004presence}.
While all three of these present their own challenges, high-quality tracking is a significant roadblock for effective \acrshort{arvr}, since errors and delay in tracking necessarily limit the performance of the subsequent steps \citep{welch1996scaat}.
This is especially true for \acrshort{ar} scenarios, where users are able to notice even small misalignments of the virtual world with the external environment.
To reduce misalignments and provide an immersive experience, there is a need of a high-frequency 6 \acrshort{dof} pose tracker.
6 \acrshort{dof} tracking is better than 3 \acrshort{dof} orientation tracking since it provides parallax cues, affords the ability to move around and much more.

The frequency of camera-based tracking systems is implicitly limited by the frame rate of the cameras.
The frame rate acts as an implicit barrier to the highest tracking frequency possible and limits the usability of cameras for tracking, where cameras often play the secondary role of drift correction, while the \hf tracking is performed by other sensors like \acrshortpl{imu} \citep{lavalle2014head}.
One of the objectives of this dissertation is to address this frame-rate barrier, enabling \acrshort{arvr} systems of the future.
I describe an approach in this dissertation by leveraging \acrfull{rs} cameras and estimating poses at frequencies as high as 80kHz.
Furthermore, I utilize radial distortion to further constrain the ego-motion, improving the system stability while maintaining accuracy and tracking frequency.

Another objective of this dissertation is to improve efficiency of optimization algorithms that enforce edge-aware constraints.
This is particularly useful for \acrshort{vr} content creation where a fast optimization scheme is required to estimate depth which is consistent with the color image \citep{anderson2016jump}.
As such systems continue to improve, higher and higher video resolutions will be necessary to meet the desired quality for immersive \acrshort{vr} experiences \citep{kanter2015graphics}.
This objective is not only limited to content creation, but useful for 3D reconstruction and texturing \citep{waechter2014let}, computational photography \citep{barron2015fast} and real-time point cloud processing \citep{valentin2018}.

\section{Dissertation Statement}
\label{sec:thesis_statement}
The frequency of pose estimation in tracking can be significantly improved by leveraging rolling shutter and radial distortion, both of which are traditionally
considered to be artifacts.
High-speed edge-aware optimization can be parallelized by leveraging approximate distance-preserving 1D filtering techniques.

\section{Outline of Contributions}
This thesis makes the following contributions that advance the state of the art in high-frequency 6 \acrfull{dof} pose estimation using rolling shutter cameras and pushes the boundaries of high-speed edge-aware optimization.
The following research has been published in the support of the research statement \citep{bapat2016towards}, \citep{bapat_rolling_radial_tracking} and \citep{bapat2018_dts}.
\paragraph{Rolling Shutter Based Tracking: }
Traditionally, \acrfull{rs} is considered to be an imaging artifact that has to be overcome and corrected \citep{forssen2010rectifying,grundmann2012calibration}.
To turn this `negative' phenomenon into a virtue, I propose to assess the feasibility of estimating a pose for each row of a rolling shutter camera in contrast to the traditional approach of estimating a pose per frame.
To that effect, I investigate whether per-row poses can be estimated assuming past image frames and motion history is available.
Additionally, I propose to develop a method for high-frequency tracking and design a multi-camera cluster of rolling shutter cameras \citep{bapat2016towards}.
I present evaluation to study the accuracy of the tracker for synthetic motion paths inside a simulator as well as human motion sequences captured using the Hi-Ball tracker~\citep{welch1999hiball}.
\paragraph{Leveraging Distortion for Tracking: }
In addition to rolling shutter, radial distortion is also considered to be an imaging artifact which needs to be corrected.
To reduce the number of cameras in the multi-camera cluster, I will leverage radial distortion.
Furthermore, I show that radial distortion in fact helps in improving the stability of the tracker \citep{bapat_rolling_radial_tracking} while maintaining requisite accuracy.
\paragraph{Edge-Aware Optimization: }
In edge-aware optimization, the goal is to optimize a set of variables,~\eg depth per pixel, while still respecting the edges in the guiding color image.
This is useful in depth refinement, optical flow estimation \citep{revaud2015epicflow} and \acrshort{vr} video stitching \citep{anderson2016jump}.
Traditionally, such optimization is conducted in a 5D space comprising of 2D pixels and 3D color.
Due to the curse of dimensionality, increasing the channels of the image makes such an approach very costly.
I propose a method which recognizes that the 5D space is sparsely populated since the image defines a function from 2D pixels to 3D color.
Using this knowledge, I propose an edge-aware optimization framework which runs at high-speed and can be applied to a variety of computer vision tasks.
I study the speed-accuracy trade-off of this efficient and parallelizable algorithm \citep{bapat2018_dts}.

\section{Dissertation Outline}
In Chapter \ref{chap:tech_intro}, I provide technical background and introduce notation used in this dissertation.
I describe the basics of imaging and camera shutter, camera models and optimization techniques used in the proposed methods.

In Chapter \ref{chap:related_work}, I discuss relevant prior work with a particular focus on \rs camera, visual odometry and efficient edge-aware optimization techniques.
I provide an overview of the \acrshort{rs} camera model and the theory developed for explaining \acrshort{rs} artifacts, \acrshort{rs} camera-based geometric estimation and calibration.
Then I describe computer vision based methods for odometry and \hf tracking. 
Finally, I review the edge-aware filters and optimization techniques and their computational complexity with a focus on high-speed and parallelizable methods.

In the papers ~\cite{bapat2016towards} and ~\cite{bapat_rolling_radial_tracking}, I develop my \hf per-row 6 \acrshort{dof} tracking framework.
I model the rolling shutter and describe how a linear motion model is useful for \hf tracking for \acrshort{arvr} applications.
Additionally, I model the radial distortion of the lens and describe how it is useful in decreasing the number of cameras in the multi-camera system, while maintaining the tracking accuracy.

In ~\cite{bapat2018_dts}, I develop the highly efficient edge-aware optimization framework and show its effectiveness for multiple tasks such as colorization, depth super-resolution and stereo.

Finally I state the limitations and outline the future research directions building upon this thesis in Chapter \ref{chap:future_work}.

\chapter{Technical introduction}
\label{chap:tech_intro}
In this chapter I will provide an overview of concepts and tools useful for understanding this dissertation.
In particular, I will introduce the traditional pinhole camera model, \acrfull{gs}, \acrfull{rs} and highlight characteristics of \acrshort{rs} camera.
Furthermore, I will discuss the lens models used in this dissertation.
This thesis uses a system linear of equations to estimate tracking, and gradient descent for edge-aware optimization.
I will briefly introduce both of these tools for parameter estimation.

\section{Image capture and pinhole cameras}
\label{sec:tech_intro:pinhole_camera}
A camera is a light capturing device, which can encode the sensed intensity either in the form of chemical changes in a film, or in the form of electric charge in \acrfull{ccd} or a \acrfull{cmos} device.
To understand the imaging process, let us follow the light during this process for a digital camera.
The light first gets emitted from a light source, reflects from the surface of the scene, passes through a lens system (if present) and bends according to the lens refraction.
When the camera opens the aperture using a shutter mechanism (physical or electronic), the light enters the camera aperture and falls on a light sensitive pixel,~\eg a \acrshort{cmos} cell.
This continues until the camera shuts the aperture, during which the camera is continuously integrating/accumulating the incident light.
Finally, the camera `reads out' the intensity values from the sensor by sequentially applying gain amplification and analog-to-digital conversion.
These intensity values are transferred to a computer and/or stored into a memory.

The imaging process for cameras in the simplest case is modeled by a pinhole camera.
Pinhole cameras (or camera obscura) are `ideal' cameras which have an infinitesimally small aperture and no lens system.
Such a camera is represented mathematically via three geometric transformations applied consecutively: \begin{enumerate*} \item pose, \item perspective division and \item scaling and re-centering according to intrinsic matrix\end{enumerate*}.
To understand these transformations, first let us define a world coordinate space.
Let us assume that the world is measured in meters and referenced using a Cartesian coordinate system.
The pinhole camera's location and orientation can be represented by a rigid transformation $\boldsymbol{V}$ also referred to as \textit{pose} as follows:
\begin{equation}
\boldsymbol{V} =   \begin{bmatrix}
  \begin{array}{c|c}
   \boldsymbol{R}_{3 x 3} & T_{3 x 1} \\
  \hline
  0_{1 x 3} & 1
  \end{array}
  \end{bmatrix},
\end{equation}
where $\boldsymbol{R}$ is the rotation matrix and $T = -\boldsymbol{R}C$, and $C$ is the location of the camera's center of projection.
The first of the transformations defined by the pose of the camera maps a 3D point $X_1$ (in meters) to another 3D point $X_2$ (in meters) in the camera space so that the world is now rotated and centered such that $\boldsymbol{R} = \boldsymbol{I}$, the identity matrix, and $T = \overrightarrow{0}$.
In other words, the camera is located at the origin with no rotation.
\begin{equation}
     \begin{bmatrix}
  \begin{array}{c}
   x_2 \\
   y_2 \\
   z_2 \\
  \hline
  1
  \end{array}
  \end{bmatrix} =  \boldsymbol{V} \begin{bmatrix}
  \begin{array}{c}
   x_1 \\
   y_1 \\
   z_1 \\
  \hline
  1
  \end{array}
  \end{bmatrix},
\end{equation}
or $X_2 = \boldsymbol{V}  X_1$ in short.
Here the 3D point $X_1 = [ x_1, y_1, z_1 ]$ is expressed in homogeneous coordinates  as $[ x_1, y_1, z_1, 1]$, similarly for $X_2$.
In the following, I will use the homogeneous representation and the non-homogeneous one interchangeably.
This is followed by the function $\pi$ which does perspective division or z-division, \ie, $\pi$ maps the 3D point (in meters) onto a 2D point (in meters) in normalized image space.
\begin{equation}
     \begin{bmatrix}
  \begin{array}{c}
   \tilde{p}_x \\
   \tilde{p}_y \\
  \hline
  1
  \end{array}
  \end{bmatrix} =  \pi\left( \begin{bmatrix}
  \begin{array}{c}
   x_2 \\
   y_2 \\
   z_2 \\
  \hline
  1
  \end{array}
  \end{bmatrix}\right),
\end{equation}
where $[\tilde{p}_x, \tilde{p}_y] = [\frac{x_2}{z_2},  \frac{y_2}{z_2}]$.
This is followed by scaling and centering of the normalized image space (in meters) into image pixel location $[p_x, p_y]$.
This last transformation is represented by the intrinsic matrix $\boldsymbol{K}_{3 x 3}$.
\begin{equation}
     \begin{bmatrix}
  \begin{array}{c}
   p_x \\
   p_y \\
  \hline
  1
  \end{array}
  \end{bmatrix} = \underbrace{\begin{bmatrix}
  \begin{array}{c c c}
   f_x & s & c_x \\
   0 & f_y & c_y \\
   0 & 0 & 1 \\
  \end{array}
  \end{bmatrix}}_{\boldsymbol{K}}
  \begin{bmatrix}
  \begin{array}{c}
   \tilde{p}_x \\
   \tilde{p}_y \\
  \hline
  1
  \end{array}
  \end{bmatrix},
\end{equation}
Ideally, $\left(c_x, c_y\right)$ is the center of the image, also known as the principal point, and $f_x$ and $f_y$ define the focal length of the camera (zoom) in $X$ and $Y$ directions respectively.
$s$ models the skew between rows and columns of the image which is usually zero for modern \acrshort{cmos} cameras.
Often, the $\boldsymbol{K}$ matrix is subsumed in the $\pi$ function as a shorthand.
\begin{figure}
    \centering
 \begin{subfigure}[t]{0.33\textwidth}
       \centering
  \includegraphics[trim={3cm 0.0cm 0.0cm 0.0cm},clip,width=\textwidth]{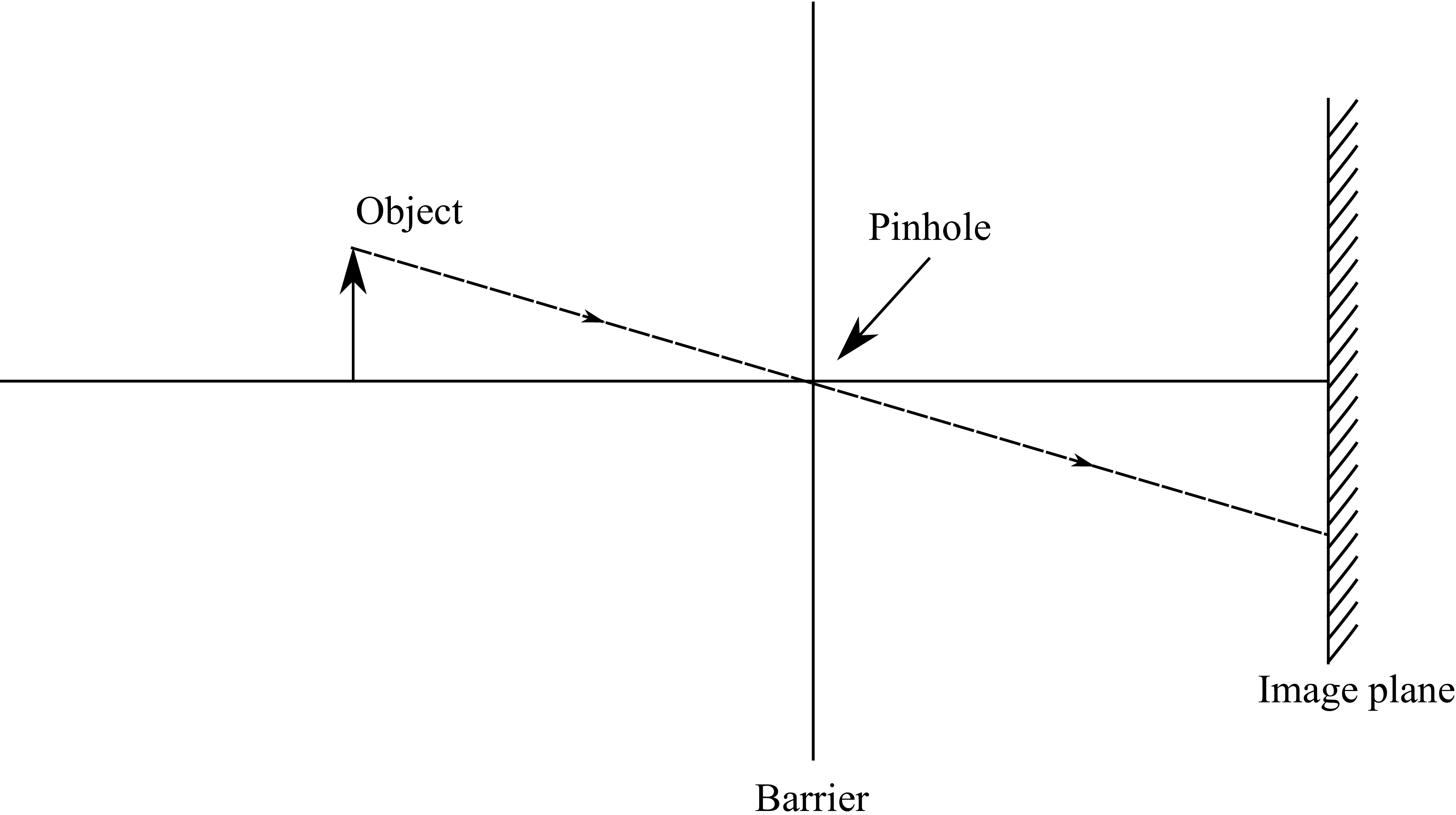}
\caption{\label{fig:pinhole_single_ray}\centering Pinhole camera, no lens.}
 \end{subfigure}%
\hfill
    \begin{subfigure}[t]{0.33\textwidth}
       \centering
    \includegraphics[trim={3cm 0.0cm 0.0cm 0.0cm},clip,width=\textwidth]{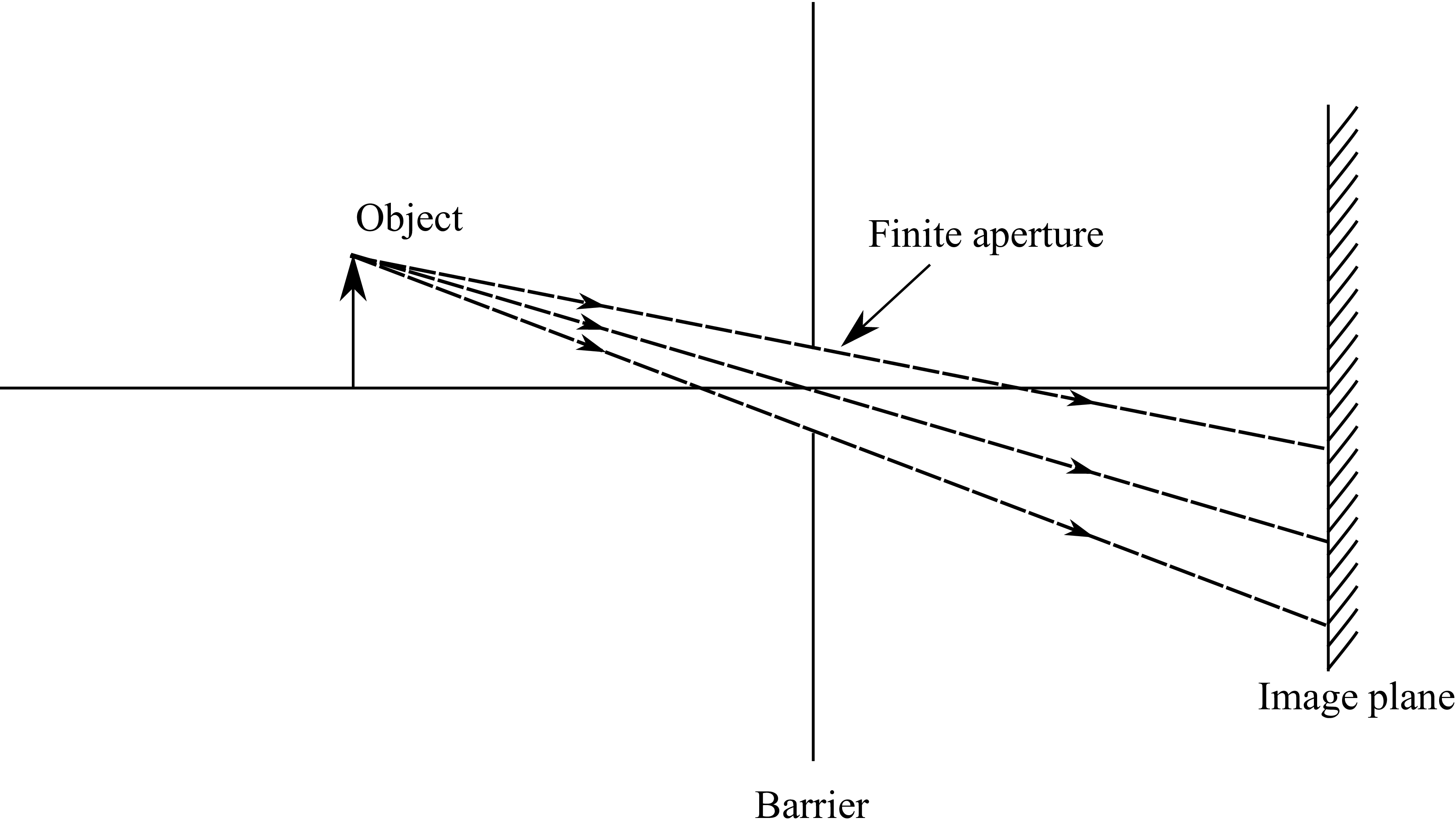}
\caption{\label{fig:finite_aperture}\centering  Camera with finite aperture, without lens.}
 \end{subfigure}%
\hfill
    \begin{subfigure}[t]{0.33\textwidth}
       \centering
    \includegraphics[trim={3cm 0.0cm 0.0cm 0.0cm},clip,width=\textwidth]{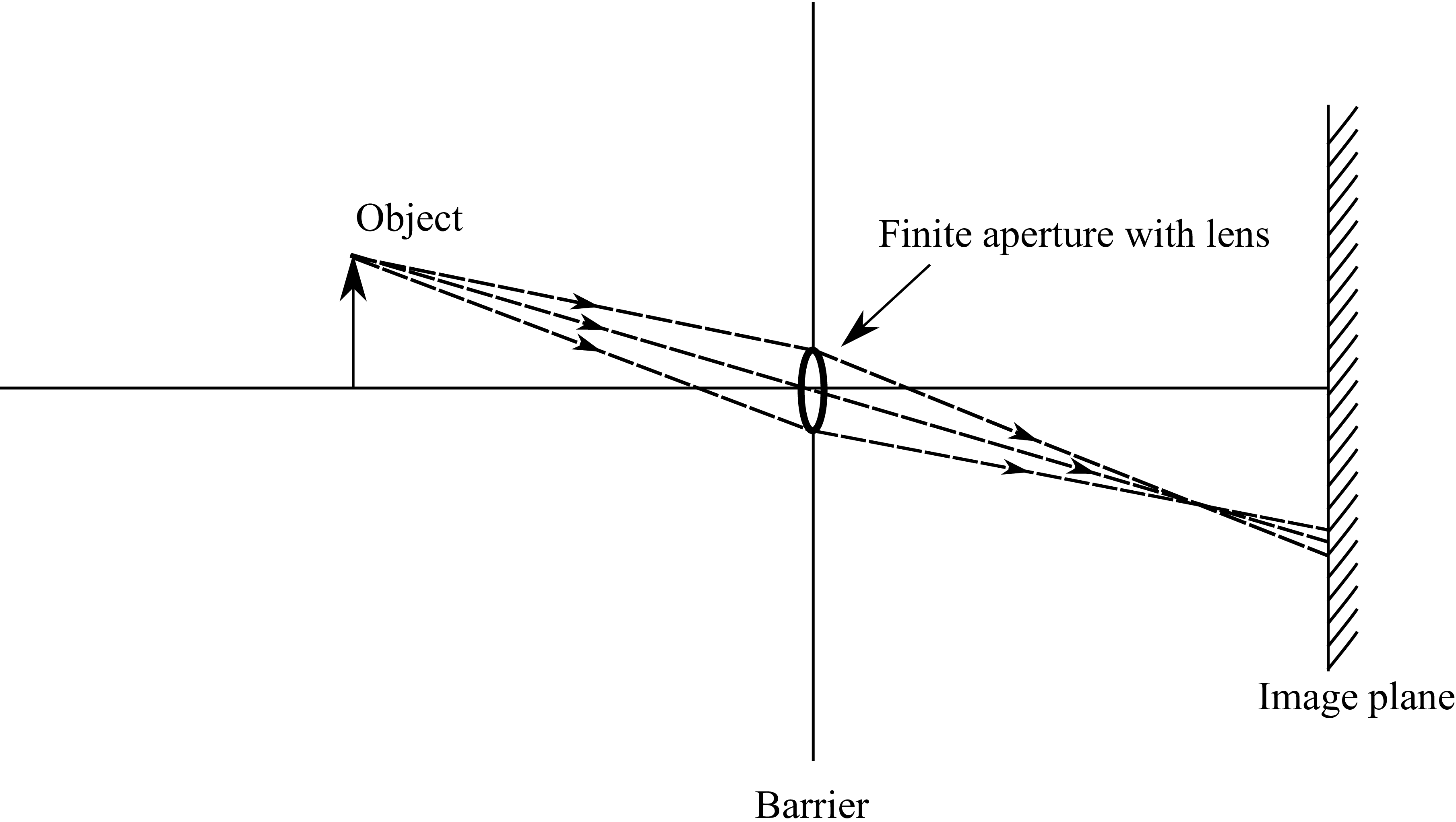}
    \caption{\label{fig:finite_aperture_w_lens}\centering  Camera with finite aperture, with lens.}
 \end{subfigure}%
\caption{\label{fig:pinhole_vs_lens}Pinhole camera with infinitesimally small aperture allows a single light ray to pass the barrier. A finite aperture allows a pencil of light rays to pass the barrier, creating a blurry image. The lens of the camera with finite aperture focuses the pencil of rays.}
\end{figure}
\section{Lens and distortion}
\begin{figure}
    \centering
 \begin{subfigure}[t]{0.32\textwidth}
       \centering
         \stackinset{l}{}{t}{}
  {\fcolorbox{red}{red}{\includegraphics[trim={20cm, 11cm, 13cm, 14cm},clip,width=0.3\textwidth]{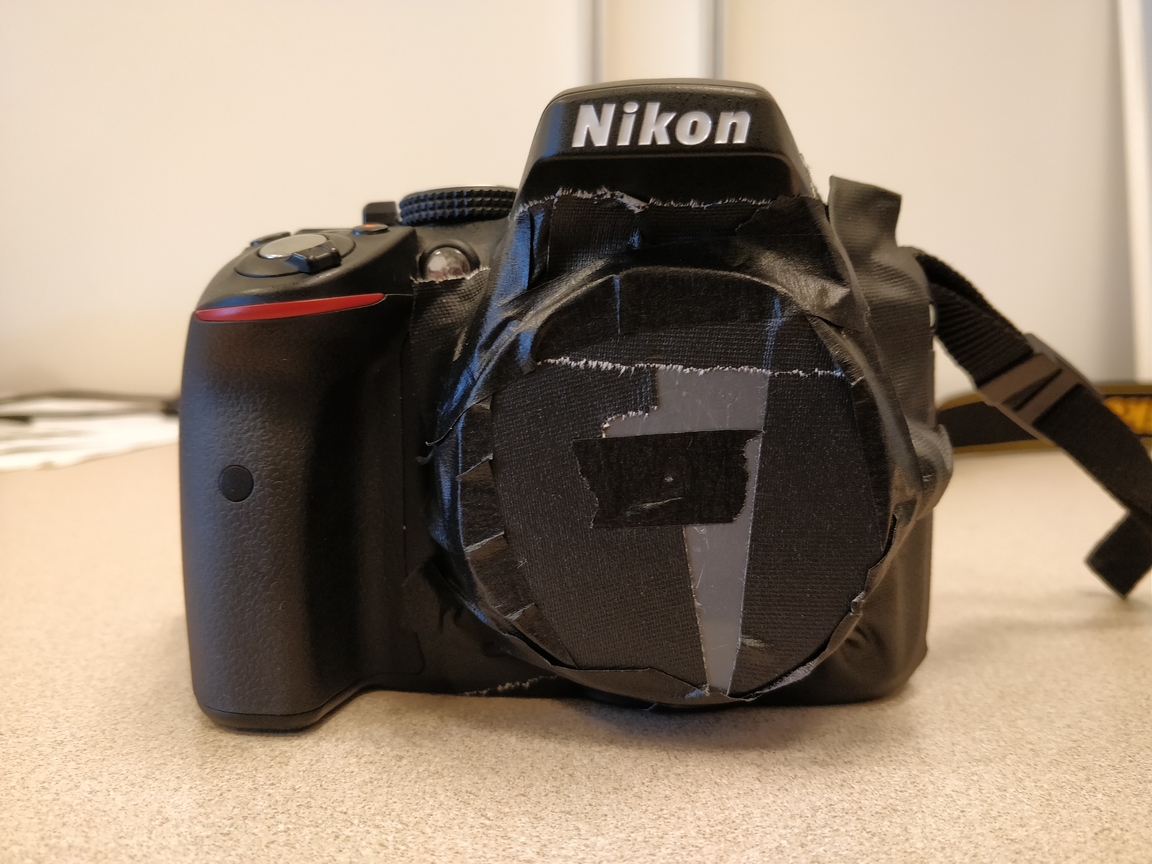}}}
  {\includegraphics[trim={0cm, 0cm, 0cm, 0.1cm},clip,width=\textwidth]{images/IMG_20190509_170445_small.jpg}}
\caption{\centering Approximate pinhole camera created from \acrshort{dslr}, the white dot in the center of inset is the pinhole.}
 \end{subfigure}%
\hfill
    \begin{subfigure}[t]{0.32\textwidth}
       \centering
    \includegraphics[trim={0.17cm 0.0cm 0.17cm 0.0cm},clip,width=\textwidth]{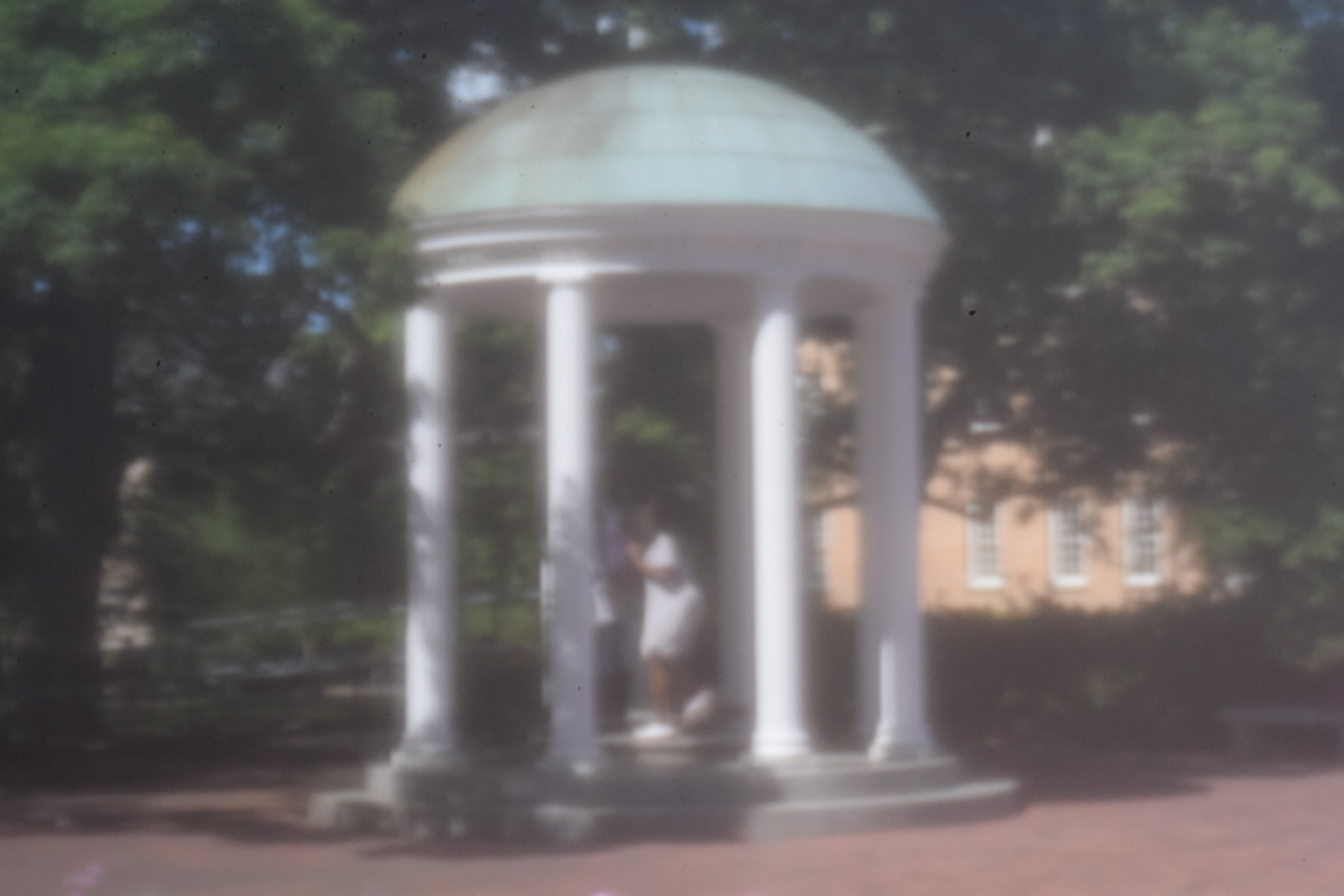}
\caption{\centering Image taken with approximate pinhole camera with 125 ms exposure time.}
 \end{subfigure}%
\hfill
    \begin{subfigure}[t]{0.32\textwidth}
       \centering
    \includegraphics[trim={0cm 0.08cm 0.1cm 0.0cm},clip,width=\textwidth]{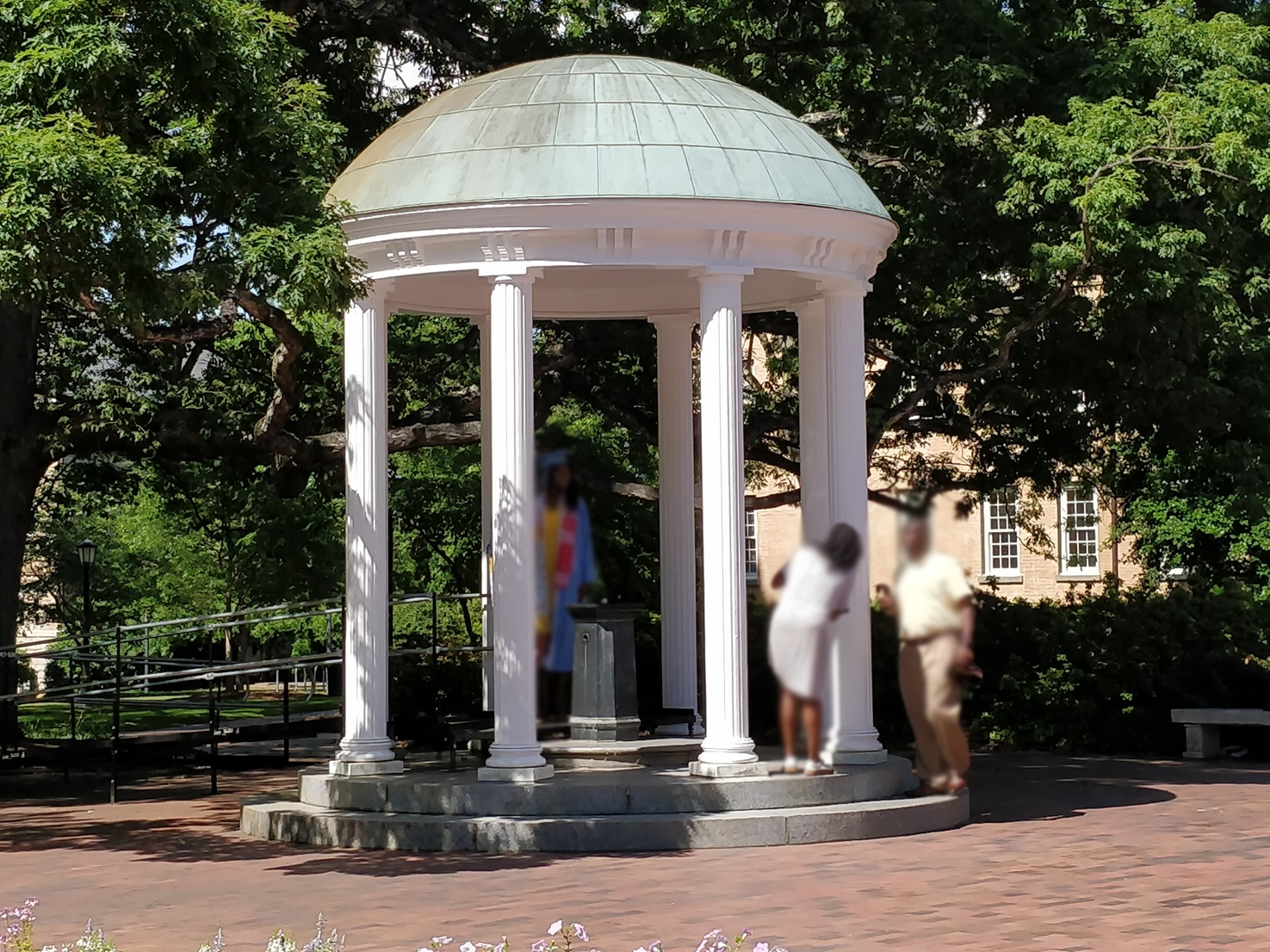}
    \caption{\centering Image taken with a OnePlus5T phone equipped with lens with 0.5 ms exposure.}
 \end{subfigure}%
    \caption{\label{fig:pinhole_vs_lens_images} Qualitatively similar images taken with a approximate pinhole camera and a OnePlus5T phone equipped with lens. As creating infinitesimally small hole is not possible, the image taken with the approximate pinhole camera appears blurry.}
\end{figure}

A pinhole camera allows a single ray to pass through the infinitesimally small aperture (hole) in the barrier as shown in Fig.~\ref{fig:pinhole_single_ray}.
The exposure length required for a pinhole camera is too long to be practical.
To collect more rays, real cameras use a finite aperture to collect more light, see Fig.~\ref{fig:finite_aperture}.
Although the finite aperture allows more light to pass, the image formed on the image plane is blurry, since the pencil of rays passing through the aperture falls on a region of the image sensor.
Real cameras employ a lens to concentrate the amount of photons incident on the camera sensor by bending the light, effectively keeping the scene in focus and reducing the exposure time, see Fig.~\ref{fig:finite_aperture_w_lens}.
Fig.~\ref{fig:pinhole_vs_lens_images} shows qualitatively similar images captured with a camera with finite aperture without any lens and OnePlus5T phone camera.
The camera with finite aperture is created using a black tape and a Nikon D5300 \acrshort{dslr} and is an approximation of a pinhole camera.
The exposure for approximate pinhole camera is 125 ms while the lens of OnePlus5T phone focuses the light onto the sensor, drastically reducing the required exposure to 0.5 ms.

Fig.~\ref{fig:finite_aperture_w_lens} depicts that the the pencil of rays converges at a single point and then diverges and falls on the image sensor.
This happens since the object in the scene is not in-focus.
This effect creates a blurry region for all out of focus objects called `circle of confusion' where the amount of blur is dependent on the depth of the object, the focal length of the lens, position of image sensor and the size of the aperture \citep{szeliski2010computer}.
Although the lens is necessary in practical cameras, it can affect the image quality due to optical aberrations such as vignetting, radial distortion, tangential distortion and color aberrations to name a few.
Out of these, I explore the radial distortion in more detail.
\subsection{Radial Distortion}
\begin{figure}
    \centering
         \begin{subfigure}[b]{0.3\textwidth}
        \centering
        \includegraphics[width=\textwidth] {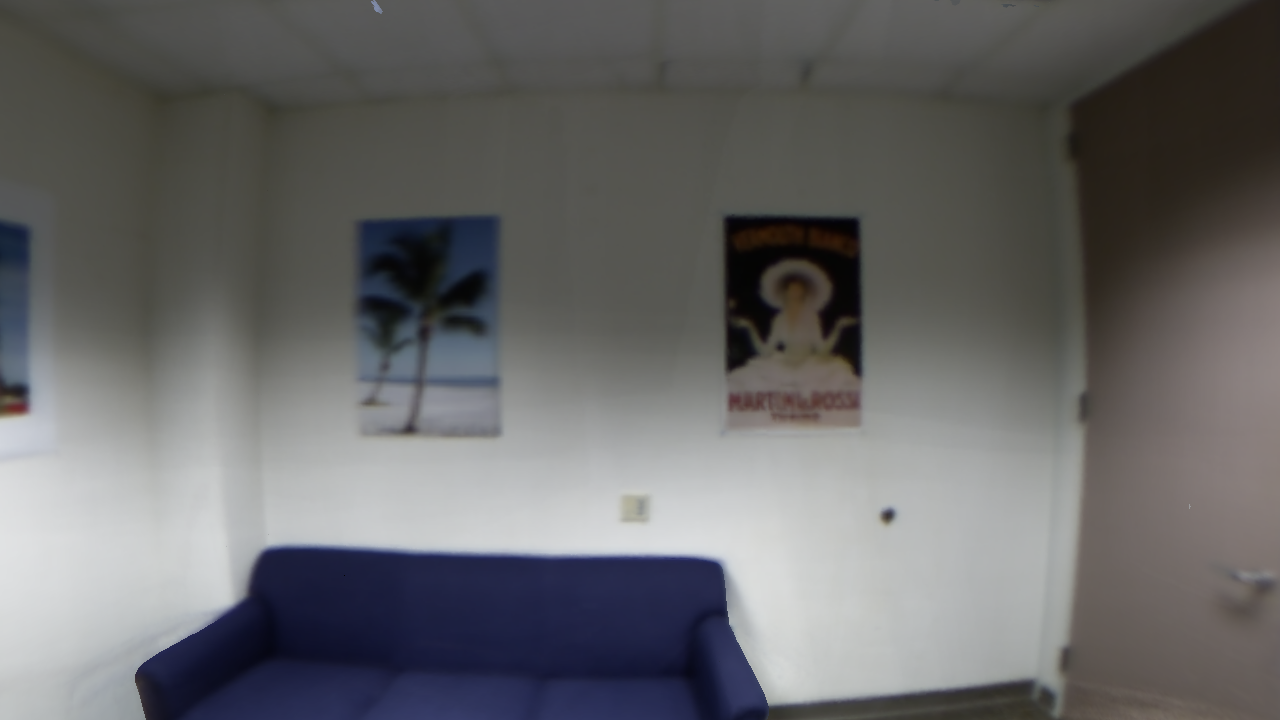}
        \caption{\label{fig:radial_disortion_barrel}Barrel distortion}
    \end{subfigure}%
 \hfill
        \centering
         \begin{subfigure}[b]{0.3\textwidth}
        \centering
        \includegraphics[width=\textwidth] {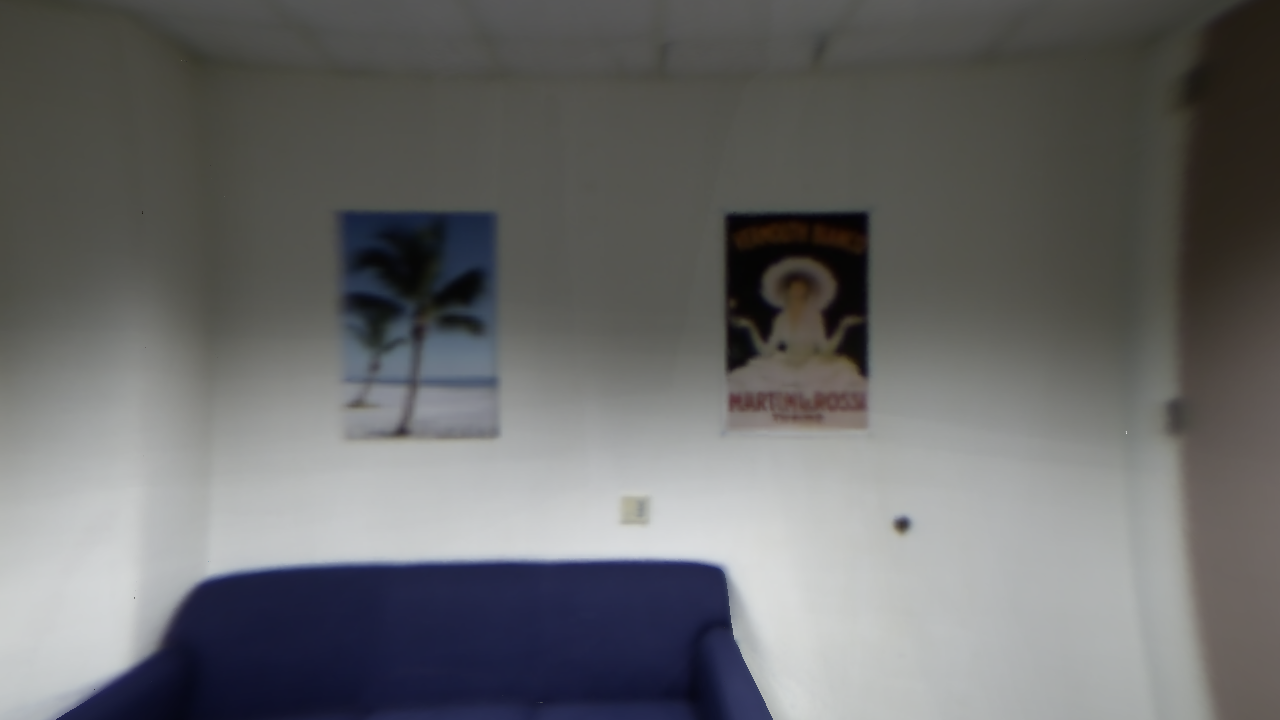}
        \caption{\label{fig:radial_disortion_pincushion}Pincushion distortion}
    \end{subfigure}%
 \hfill
      \begin{subfigure}[b]{0.3\textwidth}
        \centering
        \includegraphics[width=\textwidth] {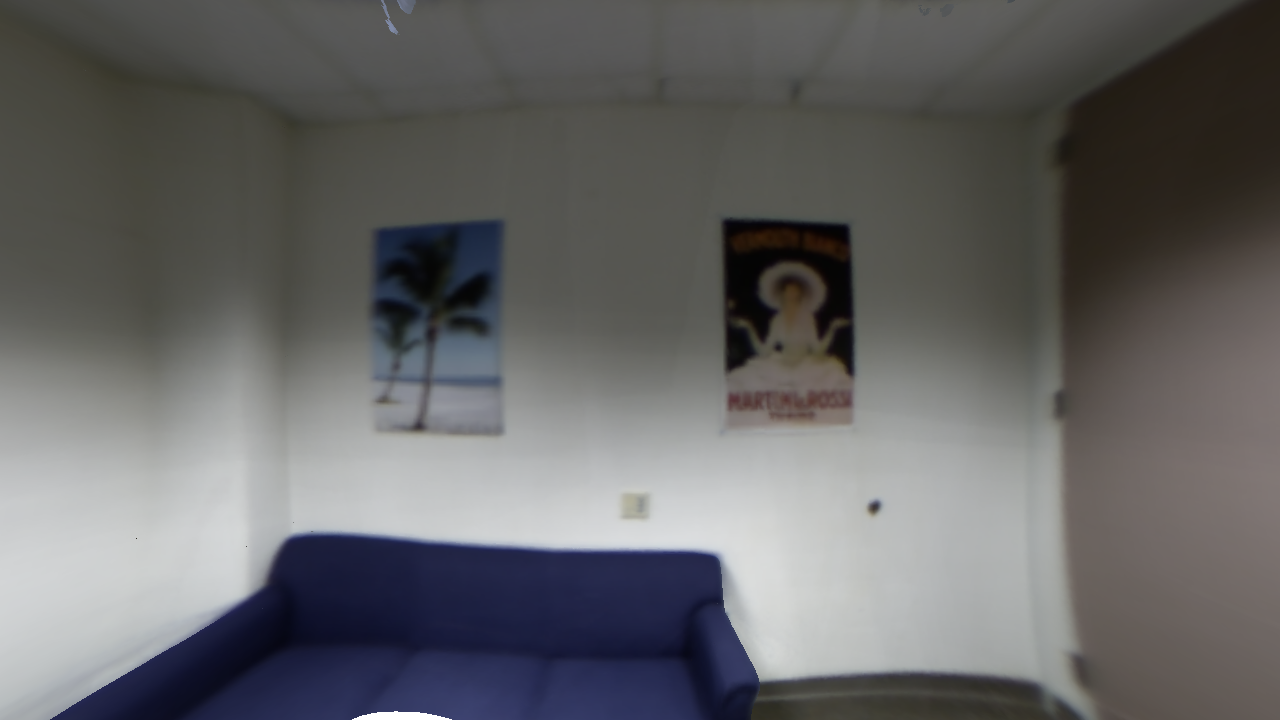}
        \caption{\label{fig:radial_disortion_moustache}Moustache distortion}
    \end{subfigure}%
    \caption{\label{fig:radial_disortion_example} Types of radial distortion.}
\end{figure}
Radial distortion manifests due to uneven magnification by the lens.
Fig.~\ref{fig:radial_disortion_barrel} depicts \textit{barrel distortion} typically observed in wide-angle lenses which tend to have higher magnification at the center than the periphery, while the inverse effect describes the \textit{pincushion distortion} as shown in Fig.~\ref{fig:radial_disortion_pincushion}.
If these distortions are simultaneously present, while less common, is called \textit{mustache distortion} due to its similarity to a handlebar mustache, see Fig.~\ref{fig:radial_disortion_moustache}.
I will refer to all of the above distortions as radial distortion.

As radial distortion is induced by the lens system, the easiest way to formulate it is to work in the normalized camera space.
Radial distortion as characterised by \cite{brown1966decentering}, also known as Brown-Conrady model~\citep{conrady1919decentred}, is described below.
The distorted normalized image space pixel location $\left(\tilde{p}_{xd}, \tilde{p}_{yd}\right)$ is modeled as a power series of the unobserved and undistorted normalized image pixel location $\left(\tilde{p}_{xu}, \tilde{p}_{yu}\right)$ using the radius $r$ as:
\begin{align}
    \tilde{p}_{xd} &= \tilde{p}_{xu} ( 1 + k_1 r^2 + k_2 r^4 +k_3 r^6 + \dotsm),\\
    \tilde{p}_{yd} &= \tilde{p}_{yu} ( 1 + k_1 r^2 + k_2 r^4 +k_3 r^6 + \dotsm),\\
    r^2 &= \tilde{p}_{xu}^2 + \tilde{p}_{yu}^2.
\end{align}
The more elaborate model described by \cite{brown1966decentering} also incorporates tangential distortion, but I forgo that here.
The distortion parameters $k_i$ along with intrinsic matrix $\boldsymbol{K}$ are estimated using camera calibration method by \cite{zhang2000flexible}.
\subsubsection{Inverse radial distortion model}
The less popular inverse radial distortion model expresses the undistorted pixel locations in terms of distorted locations.
\begin{align}
    \tilde{p}_{xu} &= \tilde{p}_{xd} ( 1 + k_1' \rho^2 + k_2' \rho^4 +k_3' \rho^6 + \dotsm),\\
    \tilde{p}_{yu} &= \tilde{p}_{yd} ( 1 + k_1' \rho^2 + k_2' \rho^4 +k_3' \rho^6 + \dotsm),\\
    \rho^2 &= \tilde{p}_{xd}^2 + \tilde{p}_{yd}^2.
\end{align}
The above power series is related to the Brown's model via inversion~\citep{tsai1987versatile}.
Although, calibration methods exist which solve for the intrinsic matrix $\boldsymbol{K}$ and $k'$ parameters of the inverse model~\citep{horn2000tsai}, in this thesis I use the Brown's model to describe radial distortion.
\section{Camera shutter}
While the lens system concentrates the incoming beam of light, the camera shutter controls the duration of integration of the light.
This duration is called the \textit{exposure time}, denoted by $t_e$.
The camera keeps sensing the light throughout the exposure time, effectively integrating the irradiance $E$ of light reaching a pixel location $p$.
The digital intensity value captured by the camera is given by
\begin{equation}
    I(p) = f_{cam}\left( \int_{0}^{t_e} E(p)\whitespace dt\right)
\end{equation}
The function $f_{cam}\left(\cdot\right)$ encapsulates the hardware and sensor characteristics particular to the camera, transforming the analog sensor response into a digital pixel intensity value.

One of the most primitive forms of shutter is a lens cap used to block light after a long exposure.
This was later replaced by the use of mechanical shutters like a diaphragm or leaf shutter.
A form of mechanical \acrfullit{rs} used one or more curtains that move across the aperture to uncover the sensor.
The following describes the electronic shutters available in modern digital cameras.

\subsection{Global shutter}
\begin{figure}
    \centering
    \includegraphics[width=\textwidth]{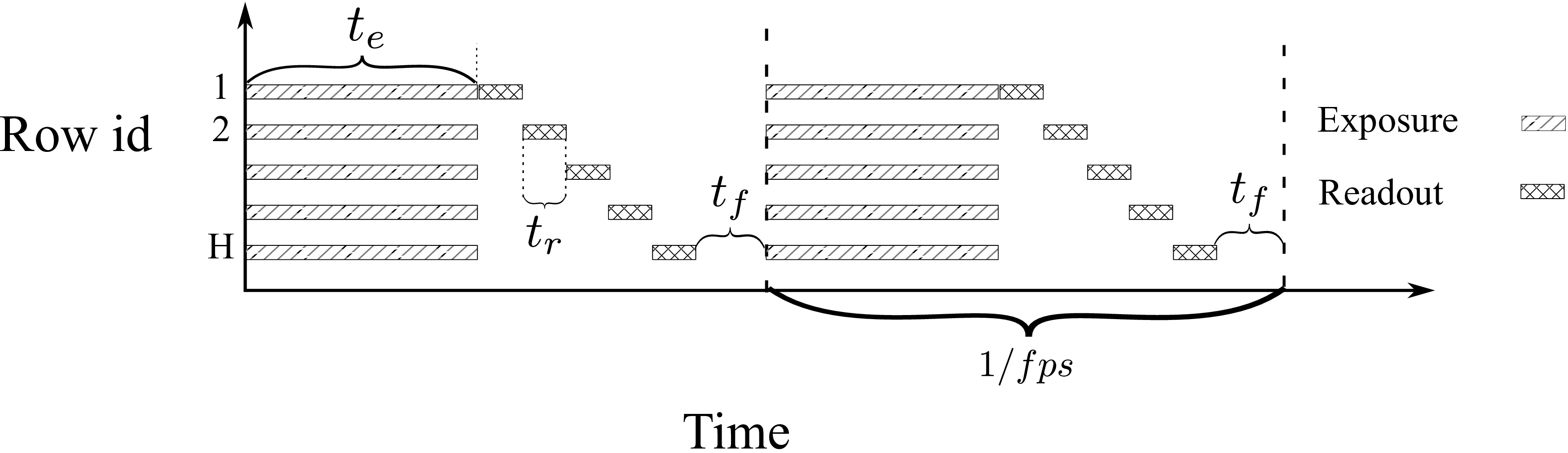}
\caption{\label{fig:gs_camera_timing}\acrshort{gs} camera timing characteristics: All the rows are exposed for $t_e$ time simultaneously, and the readout of consecutive rows is staggered by $t_r$. After the $H$ rows of the frame are exposed and readout, the capture for the next frame starts after $t_f$ time.}
\end{figure}

Traditional film cameras, as well as \acrshort{ccd} arrays in early digital cameras, exposed the entire 2D image space during a common exposure period $t_e$; this type of exposure is called global shutter.
As digital cameras became popular, electronic \acrfullit{gs} became the defacto standard.
After a digital camera stops exposure, the readout hardware starts extracting the grayscale values from each row sequentially spending $t_r$ time per row, see Fig.~\ref{fig:gs_camera_timing}.
After readout, the camera spends $t_f$ time called frame-delay to reset the pixels and prepare for the next frame.
For some \acrshort{gs} cameras, $t_f$ can be zero, or the readout is done simultaneously to support higher \acrfull{fps}.

To support simultaneous exposure, each pixel in a \acrshort{gs} camera has an additional local memory~\citep{ge2012design, witters20031024x1280}.
After the end of exposure time $t_e$, the photo-generated charges are transferred to the local memory. 
This results into smaller photo-sensitive area in the pixel, thus reducing the amount of light gathered.
As each pixel has the local memory, this effectively increases the cost and at the same time limits the resolution of the sensor by physically limiting the packing density of the pixels~\citep{ge2012design}.
Affordable commodity cameras avoid this by time-multiplexing the readout circuitry and exposure time by employing a \acrfullit{rs}, which we will look at next.
\subsection{Rolling shutter}
\label{sec:tech_intro:rs}
\begin{figure}
    \centering
    \begin{subfigure}[b]{0.35\textwidth}
    \centering
    \includegraphics[width=0.95\textwidth,height=7.725cm]{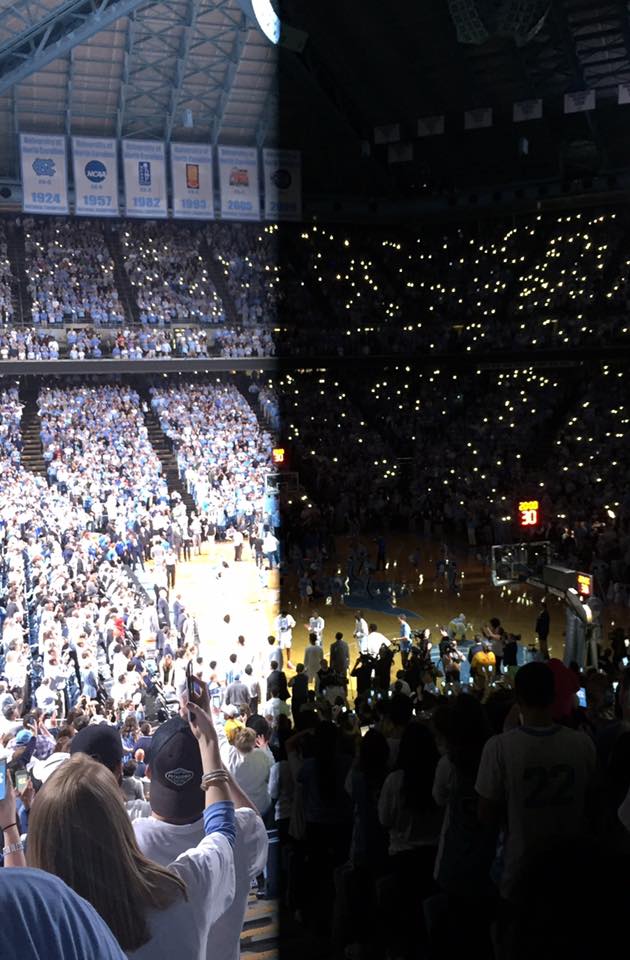}
    \caption{\label{fig:rs_light_switch_artifact}The lights switched off while the \acrlong{rs} camera was exposing the frame creating a half dark and half bright image. Image courtesy of Chelsey McCotter.}
        \end{subfigure}\qquad
           \centering
         \begin{subfigure}[b]{0.3\textwidth}
        \centering
        \includegraphics[trim={1.1cm 0.0cm 0.8cm 0.5cm},clip,width=\textwidth,height=4cm] {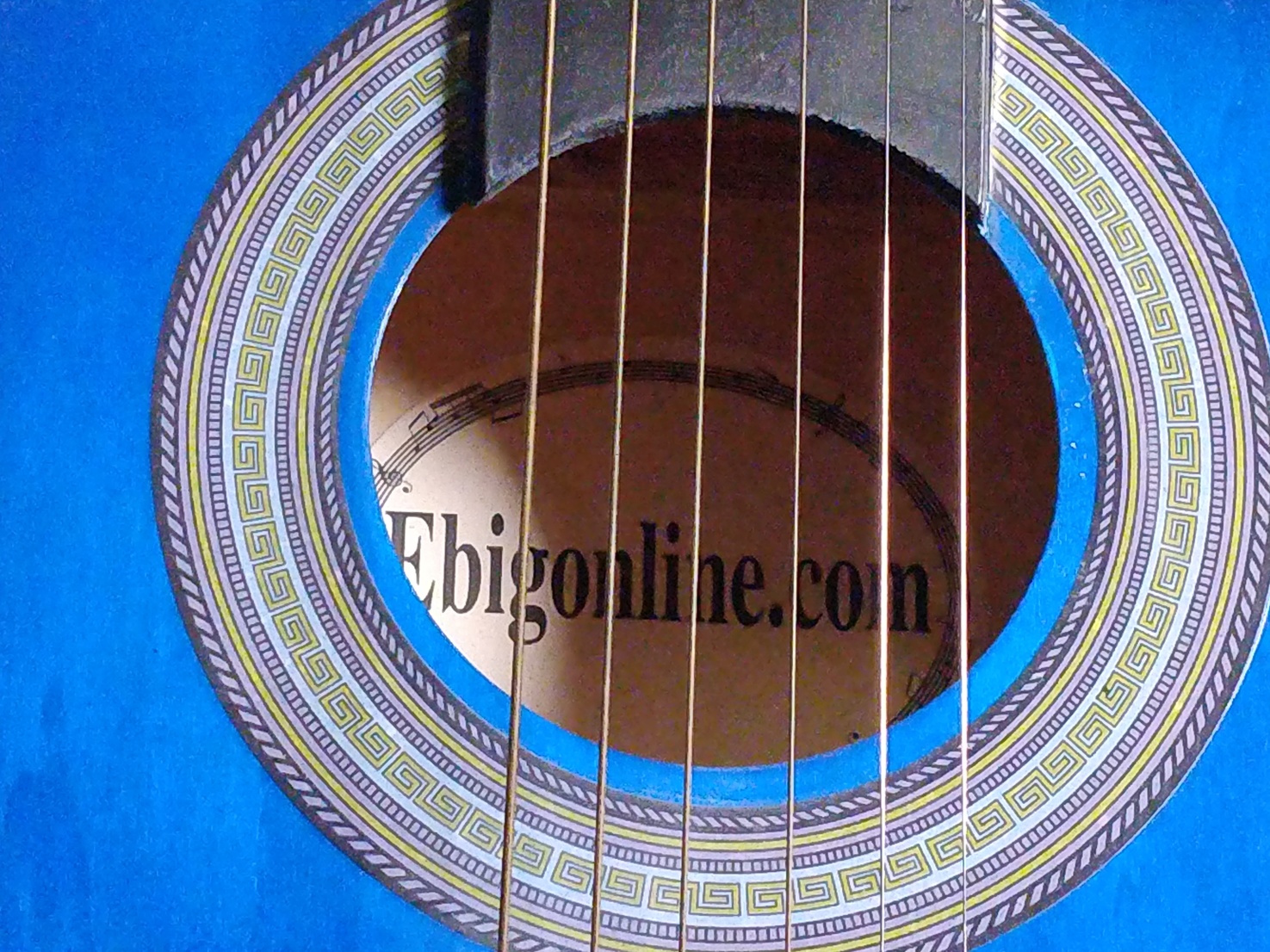}
        \caption{\label{fig:rs_static_guitar}Static guitar strings.}
        \vspace{1ex}
        \centering
        \includegraphics[trim={1.2cm 0.0cm 0.7cm 0.4cm},clip,width=\textwidth,height=4.2cm] {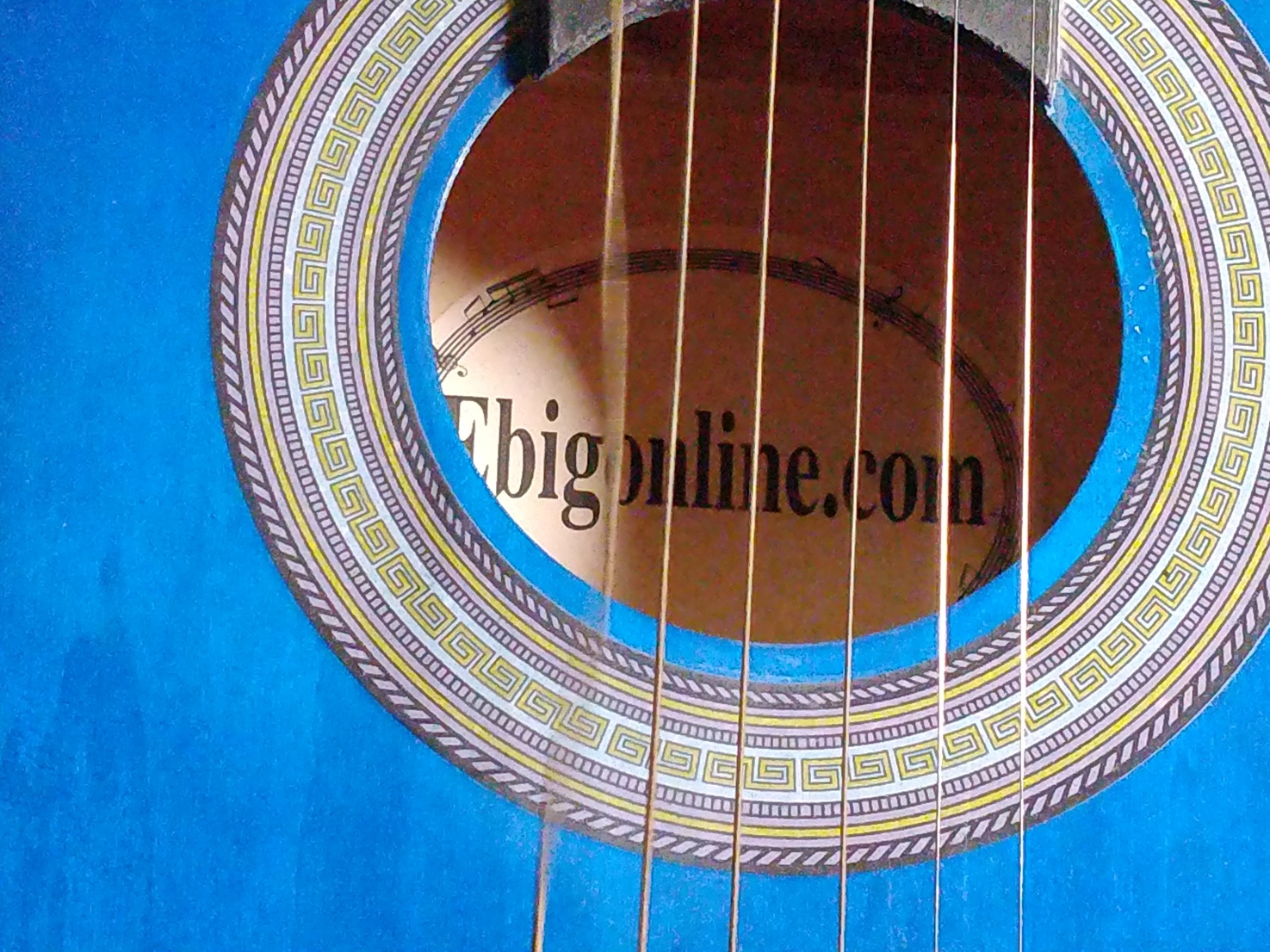}
        \caption{\label{fig:rs_guitar_artifact}\Acrlong{rs} causes the string to split after plucking.}
    \end{subfigure}%
\caption{\label{fig:rs_camera_artifact} \Acrlong{rs} artifacts: As the \acrlong{rs} sweeps in the horizontal direction, artifacts are induced.}
\end{figure}

With the advent of low-power \acrshort{cmos} camera sensors, \acrfull{rs} cameras have quickly become ubiquitous for everyday camera-equipped devices.
To save space on the silicon chip, the newer \acrshort{cmos} sensors perform sequential exposure of sensor rows, which only requires a single, simpler set of readout circuitry to read all rows and minimal buffer memory \citep{bradley2009synchronization}.
In this fashion, each individual row of the \rs image is actually a snapshot of the scene at a different time period.
For image shots without significant scene or camera motion, rolling shutter is often an effective, low-cost imaging technique.
However, when the scene is dynamic, \eg strings of a guitar (Fig.~\ref{fig:rs_guitar_artifact}), or sudden changes in ambient light (Fig.~\ref{fig:rs_light_switch_artifact}), or when the camera has fast motion relative to the frame rate of the camera, artifacts such as wobble, skew and other undesirable effects become apparent \citep{forssen2010rectifying}.

\begin{figure}
    \centering
    \includegraphics[width=\textwidth]{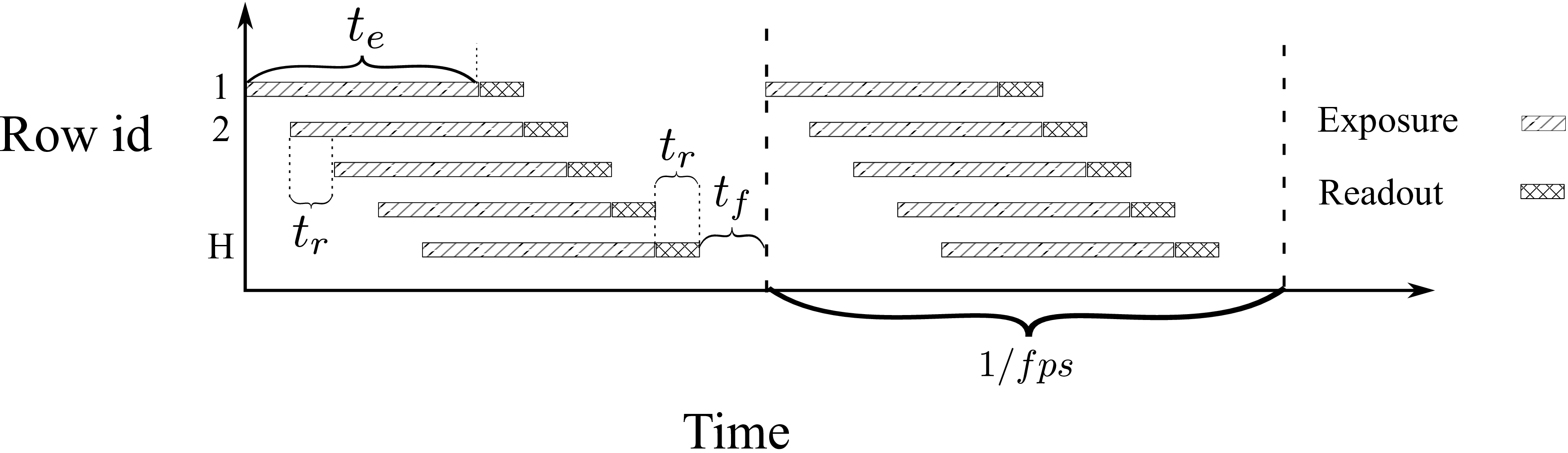}
\caption{\label{fig:rs_camera_timing}\acrshort{rs} camera timing characteristics: Each row is exposed for $t_e$ time, where the integration between consecutive rows is staggered by $t_r$. After the $H$ rows of the frame are exposed and readout, the capture for the next frame starts after $t_f$ time.}
\end{figure}

A \acrshort{rs} camera can be characterized by the exposure time $t_e$, the row (line) time delay $t_r$ between the start of exposure of consecutive rows and $t_f$, the frame delay between consecutive frames.
The camera sensor exposes a row for $t_e$ time and at the end of each row exposure, consumes $t_r$ time to read it out, see Fig.~\ref{fig:rs_camera_timing}.
As this process time-multiplexes the readout circuitry across all rows, the start of consecutive row exposures is staggered by $t_r$.
The time between the end of the readout of the last row and the start of the exposure of first row is typically used for noise reduction and reset \citep{vsmid2017rolling}. 
In general, all of these are related to the \acrshort{fps} of the camera (see Fig.~\ref{fig:rs_camera_timing}):
\begin{equation}
\frac{1}{\mathrm{fps}} = H t_r + t_f + t_e,
\end{equation}
where the rolling shutter is sweeping from top to bottom of the image and $H$ is the image height.
For a starting time $\tau_0$, the $y^{th}$ row of the camera of frame id $f_i$ in Fig.~\ref{fig:rs_camera_timing} is exposed at time $t$ given by:
\begin{equation}
     t = \tau_0 + y t_r  + f_i\whitespace\frac{1}{\mathrm{fps}}.
 \label{eq:tech_intro:row_exposure_time}
\end{equation}
\subsubsection{\glsentrytext{rs} camera model}
In rolling shutter capture, each row (or column) is captured at a slightly different time.
For simplicity, let us consider that the rolling shutter sweeps from top to bottom of the image frame.
To correctly model rolling shutter, one has to consider a (potentially) different pose for each row in the image.
To model the rolling shutter effect, let us modify the pinhole camera model from Section \ref{sec:tech_intro:pinhole_camera}.
The pinhole camera model projects the 3D point $X$ according to the pose $\boldsymbol{V}$ onto the pixel $[ p_x, p_y, 1]^T$ via the perspective projection function $\pi$.
In the following the intrinsic matrix is subsumed in the function $\pi$.
For a \rs camera, the pose $\boldsymbol{V}$ changes with time, and the camera projection can be written similarly as:
\begin{equation}
  \begin{bmatrix}
  \begin{array}{c}
   p_x(t) \\
   p_y(t) \\
  \hline
  1
  \end{array}
  \end{bmatrix}  = \pi \left(\boldsymbol{V}(t)\whitespace X \right)
\end{equation}
Remember that the Y coordinate $p_y(t)$ is analogous to the time according to Eqn.~\ref{eq:tech_intro:row_exposure_time}.
\section{Optimization}
Mathematical optimization is a technique of parameter estimation given a maximization or minimization problem.
In the minimization setting, given a function $f_{opt}(\Theta,\mathcal{D})$ of observed data $\mathcal{D}$ and unknown parameters $\Theta$, we want to estimate $\Theta$ such that $f_{opt}(\cdot)$ is minimized.
This general framework of optimization encapsulates two methods we are interested in here:
\begin{enumerate*}
\item gradient descent, and
\item the heavy-ball method.
\end{enumerate*}
\subsection{Gradient descent}
\begin{figure}
    \centering
    \includegraphics[width=0.65\textwidth]{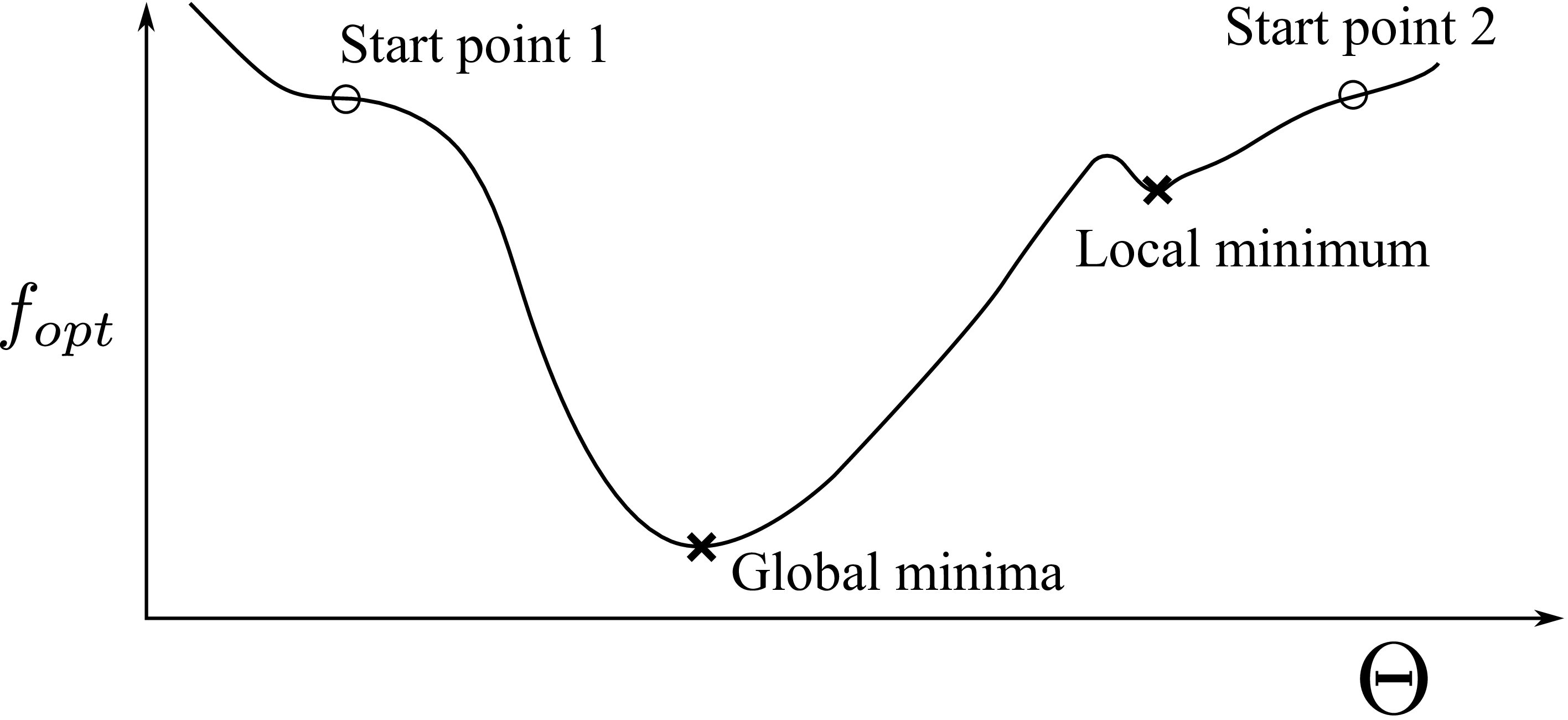}
\caption{\label{fig:gradient_descent} Gradient descent algorithm gets stuck at local minimum while the Heavy-Ball algorithm crosses the hump and converges to global minima when starting at point 2. When starting at point 1, both the update schemes converge to global minima.}
\end{figure}

Gradient descent, also known as steepest gradient descent, is an iterative algorithm which estimates the parameters $\Theta$ at every iteration by using the local gradient direction $\nabla f_{opt}\left(\Theta\right)$.
To estimate the $\Theta_{n}$ at the $n^{th}$ iteration, gradient descent defines the following update rule:
\begin{equation}
    \Theta_{n} = \Theta_{n - 1} - \gamma \nabla f_{opt}\left(\Theta\right),
\end{equation}
where $\gamma$ is the step size.
For more details on estimating optimal $\gamma$ and convergence rate of gradient descent, I refer the reader to \cite{nocedal2006numerical}.
Gradient descent will give the optimal solution for a convex function, but converges to a locally optimal solution for general functions.
For example, in Fig.~\ref{fig:gradient_descent}, when the starting point $\Theta_{0}$ is `start point 1', gradient descent will converge to the global minima, but if $\Theta_{0}$ is `starting point 2', then it will get stuck at the local minima due to the bump ahead.

\subsection{Heavy-Ball method}
Heavy-Ball method by \cite{polyak1964some} is a flavor of gradient descent which uses the past gradient updates by maintaining the \textit{momentum}.
It is inspired by the second order \acrfull{ode} which models the motion of a ball under friction $f_{opt}$ and a potential field:
\begin{equation}
\ddot{\Theta} + a \dot{\Theta} + b \nabla f_{opt}\left(\Theta\right) = 0
\end{equation}
The discrete version of this \acrshort{ode} is
\begin{equation}
\Theta_{n} = \Theta_{n - 1} - \gamma \nabla f_{opt} + \beta \underbrace{\left( \Theta_{n - 1} - \Theta_{n - 2}\right)}_{\text{momentum}}
\label{eq:heavy_ball_update}
\end{equation}
When $\beta = 0$, this method is the same as gradient descent.
The momentum term in Eq.~\ref{eq:heavy_ball_update} allows the optimizer to cross small bumps in the cost function as in Fig.~\ref{fig:gradient_descent} when $\Theta_{0}$ is `start point 2'.
Hence the Heavy-Ball method in this particular case, but not all, would converge to the global minima.
Additionally, the momentum reduces jitter in the gradient by effectively smoothing it when traversing along the cost surface during minimization.
\section{Solving a system of linear equations}
A set of linear constraints on the unknown parameters defines a system of linear equations.
Here, I will focus on the non-homogeneous case of system linear of equations defined by matrix $\boldsymbol{A}$ and vector $B$.
If the matrix $\boldsymbol{A}$ is a full-rank square matrix, then the $\Theta$ can be directly computed using the inverse.
\begin{align}
 \boldsymbol{A} \Theta  = B\\ 
   \Theta = \boldsymbol{A^{-1}} B \nonumber
\end{align}
This thesis utilizes an over-constrained system of equations, \ie more linear equations are available than the number of unknowns in $\Theta$.
Alternatively, $\boldsymbol{A}$ is a full-rank non-square matrix.
All the available constraints cannot be satisfied exactly (in general) and hence we seek a least-squares solution.
The solution to such an over-constrained system can be obtained using a number of methods such as Moore-Penrose pseudo-inverse, Cholesky decomposition and \acrfull{svd}.
I will refer the interested reader to \cite{press2007numerical} for details about these methods.
Now I will describe how the linear system's stability is characterized by analyzing $\boldsymbol{A}$ and $B$.

\subsection{Condition Number}
The estimation of $\Theta$ for a square matrix $\boldsymbol{A}$ can be reformulated as a minimization problem:
\begin{equation}
  \min_{\Theta} || \boldsymbol{A} \Theta  - B ||_{2}.
\end{equation}
For an error $\delta b$ in $B$, the relative ratio of change in solution $\Theta$ to the relative change in $B$ is
\begin{equation}
   \kappa(\boldsymbol{A}) =\max\left( \frac{\frac{|| \boldsymbol{A}^{-1} \delta b||}{||\boldsymbol{A}^{-1} B||}}{\frac{||\delta b||}{||B||}}\right) = \max\left( \frac{|| \boldsymbol{A}^{-1} \delta b||}{||\delta b||} \frac{|| \boldsymbol{A}  \Theta||}{||\Theta||}\right) = ||\boldsymbol{A}^{-1}||\whitespace||\boldsymbol{A}|| \ge 1
\end{equation}
$\kappa(\boldsymbol{A})$ is also know as the condition number of the linear system, and for $l_{2}$ norm, it can be expressed as the ratio of singular values $\sigma$ of $\boldsymbol{A}$:
\begin{equation}
   \kappa(\boldsymbol{A}) =\frac{\sigma_{\max}(\boldsymbol{A})}{\sigma_{\min}(\boldsymbol{A})} \ge 1
\end{equation}
If the system is ill-conditioned, $\kappa(\boldsymbol{A}) \gg 1$.
Analyzing the condition number is useful for checking the system stability, and I will use it to indirectly compare multi-camera tracking systems in ~\cite{bapat_rolling_radial_tracking}. 
Weighted system of linear of equations places a weight per equation by using a diagonal matrix $\boldsymbol{W}$.
The optimization reformulation can be rewritten as follows:
\begin{equation}
      \min_{\Theta} ||\boldsymbol{W} \left( \boldsymbol{A} \Theta  - B \right)||_{2}
      \label{eq:wls_definition}
\end{equation}
The weights $\boldsymbol{W}$ are carefully chosen such that the condition for the system in Eq.~\ref{eq:wls_definition} $\kappa(\boldsymbol{WA})$ decreases.

\chapter{Related Work}
\label{chap:related_work}
Rolling shutter camera, visual tracking and edge-aware optimization each have been studied extensively.
The following sections describe the relevant computer vision techniques and recent advances in these topics.
\section{Rolling shutter}
\label{sec:related_work:rs}
The ubiquitous presence of \acrfull{rs} cameras in almost every cell-phone and low-cost camera has driven much research to \begin{enumerate*} \item simplify the \rs camera model by limiting motion, \item to remove \rs induced artifacts, \item to adapt traditional 3D methods for \rs cameras, and \item to estimate the timing parameters to calibrate the camera.
\end{enumerate*}
Next, I will discuss each of these in detail.
\subsection{Motion models}
The rolling shutter projection of a 3D  point $X$ according to the time-varying pose $\boldsymbol{V}(t)$ onto the pixel $[ p_x(t), p_y(t), 1]^T$ is given by:
\begin{equation}
  \begin{bmatrix}
  \begin{array}{c}
   p_x(t) \\
   p_y(t) \\
  \hline
  1
  \end{array}
  \end{bmatrix}  = \pi \left(\boldsymbol{V}(t)\whitespace X \right)
\end{equation}
The time-varying pose $\boldsymbol{V}(t)$ is often unknown, hence removing \rs induced artifacts or directly applying traditional 3D methods for \rs cameras becomes difficult.
In general, estimating the pose $\boldsymbol{V}(t)$ of a row at a particular time $t_{p}$ is also not possible, as enough information is not present in a single row.
Hence, a variety of motion models have been explored to simplify the \rs camera projection model.
Here, I will dive into more detail for a select few.
In the following the intrinsic matrix is subsumed in the function $\pi$.
\myparagraph{Translation only model}
Translation only model assumes that the \rs camera undergoes translation with a constant orientation during the image capture.
This is especially true for \acrfull{adas} \citep{saurer2015minimal}, or for systems with high linear (translational) velocity relative to angular (rotational) velocity.
For such a motion model, the \rs reprojection can be simplified to the following:
\begin{equation}
  \begin{bmatrix}
  \begin{array}{c}
   p_x(t) \\
   p_y(t) \\
  \hline
  1
  \end{array}
  \end{bmatrix}  = \pi \left(\begin{bmatrix}
  \begin{array}{c|c}
  \boldsymbol{R_{0}} & T(t) \\
  \hline
  0 & 1
  \end{array}
  \end{bmatrix} X \right),
\end{equation}
where $\boldsymbol{R_{0}}$ is the initial rotation of the camera, which is often assumed to be the identity matrix.
Two models are used to put further constraints on the intra-frame motion: \begin{enumerate*} \item uniform (constant) velocity, and \item constant acceleration\end{enumerate*}.
These constraints further limit the complexity of the model and make the computation tractable \citep{purkait2018minimal}.
Polynomial models which capture more complex motion are also used to model intra-frame translation \citep{liang2008analysis}.
\myparagraph{Rotation only model}
A rotation only model is effective to model hand-held device motion, and is often used to correct distortions induced by \rs in smartphones.
Information about angular acceleration and angular velocity can be obtained using external sensors like \acrshortpl{imu} which provide priors on the motion, simplifying the motion estimation \citep{lee2018calibration}.
This rotation only model is defined as follows:
\begin{equation}
  \begin{bmatrix}
  \begin{array}{c}
   p_x(t) \\
   p_y(t) \\
  \hline
  1
  \end{array}
  \end{bmatrix}  = \pi \left(\begin{bmatrix}
  \begin{array}{c|c}
  \boldsymbol{R}(t) &  T_0 \\
  \hline
  0 & 1
  \end{array}
  \end{bmatrix} X \right),
\end{equation}
where $T_0$ is the initial camera position, and is often assumed to be the zero vector ($\overrightarrow{0}$).
\cite{hanning2011stabilizing} use an \acrfull{ekf} based method to filter the \acrshort{imu} measurements to estimate rotation and perform rolling shutter compensation and video stabilization.
These methods require calibration between the camera and the \acrshort{imu}.
\cite{karpenko2011digital} presented such a calibration method which uses a single video with a rotation only motion model to estimate the calibration and correct for \rs distortions.
\subsection{Removing \glsentrytext{rs} artifacts}
\begin{figure}
    \centering
         \begin{subfigure}[t]{0.3267\textwidth}
        \centering
        \includegraphics[width=\textwidth,height=5cm] {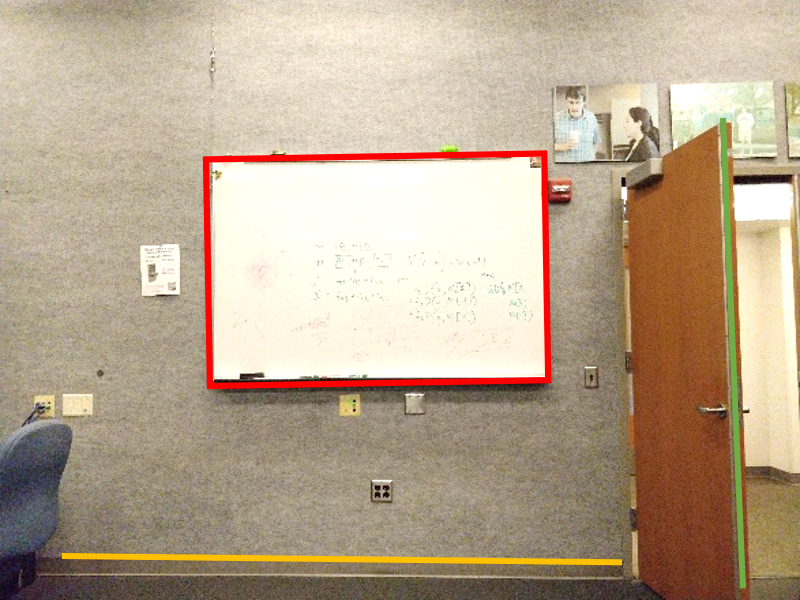}
        \caption{Static RS camera.}
    \end{subfigure}%
 \hfill
         \begin{subfigure}[t]{0.66\textwidth}
        \centering
        \includegraphics[width=0.495\textwidth,height=5.1cm] {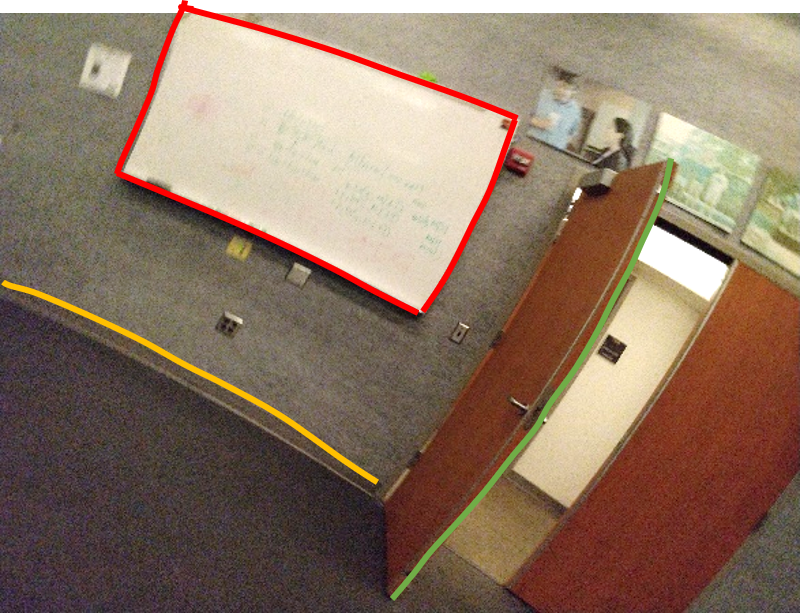}
\hfill
        \includegraphics[trim={0cm 1cm 1cm 2cm},clip,width=0.495\textwidth,height=5cm] {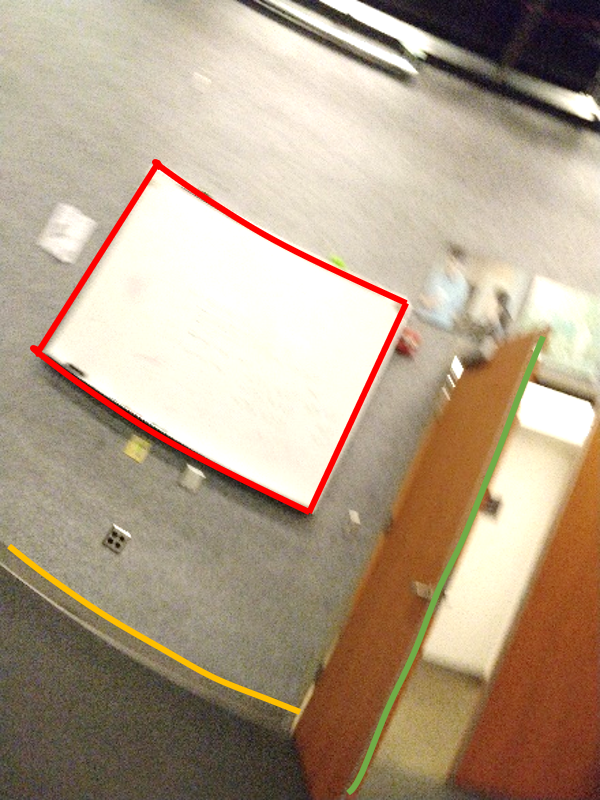}
        \caption{Dynamic RS camera.}
    \end{subfigure}%
    \caption{\label{fig:rs_artifacts} RS image captured with a static camera is the same as a GS image. Rolling shutter in the present of motion on the other hand induces different artifacts depending upon motion of the camera and the structure/depth scene.}
\end{figure}
Traditionally rolling shutter is regarded as a `distortion' introduced by the capture procedure of the cameras.
Fig.~\ref{fig:rs_artifacts} shows images taken with OnePlus5T camera, with and without motion.
\rs effect squishes, skews and bends the straight lines in the scene, depending upon the motion of the camera and the structure/depth of the scene.
Even if the scene is known beforehand, removing these artifacts is impossible without knowing a per-row pose, since different motions induce different artifacts.
Researchers have long striven to remove rolling shutter and the artifacts resulting from camera motion and the different exposure start times of the various rows of the camera have generally been regarded as nuisances of the imaging process.
\rs artifact removal has been studied extensively, and I will refer the interested reader to a rich set of approaches here \citep{liang2008analysis, heflin2010correcting, forssen2010rectifying, GrundmannKwatra2012, purkait2017rolling, rengarajan2016bows}.
\cite{ringaby2012efficient} parametrized intra-frame rotation with a linear spline and used this model for image rectification and video stabilization.
In their approach, homographies between image rows are used to reconstruct the rectified image.
In ~\cite{bapat2016towards}, I utilize a similar approach and leverage such per-row homographies to compensate for rotation in the \rs image.
More recently, \cite{rengarajan2017unrolling} proposed to correct the artifacts by utilizing the widely popular \acrfullpl{cnn} to predict polynomial models of rotation in Z direction and translation in X.
They argue that these motion pattern induce the most noticeable \rs artifacts.

In addition to \rs induced artifacts, the combination of large exposure and strong motion induces motion blur aggravating the distortions.
\cite{su2015rolling} proposed a method to remove motion blur using a single rolling shutter image.
Their method explicitly models for \rs to estimate the motion during exposure by fitting polynomials to each degree of freedom of camera motion.
Using this estimate, they invert the motion induced blur.
\cite{meilland2013unified} demonstrate a real-time \acrfull{sam} system for \acrshort{rgbd} sensor while accounting for motion blur in a \rs camera by assuming uniform velocity over the image.

\subsection{Geometric methods for \glsentrytext{rs}}
The recent popularity of \rs cameras in smartphones has led to a trend where traditional geometric algorithms are being adapted to model \rs.
The following explores efforts to address the absolute pose problem, the relative pose problem, stereo, \acrfull{sfm} and calibration algorithms for \rs cameras.

\myparagraph{Minimal Gr{\"o}bner basis solvers for {PnP}}
The \acrfull{pnp} problem seeks to estimate the pose of a calibrated camera given $n$ correspondences between 3D points and their 2D pixel projections in the image.
Minimal solvers sample the minimal number of correspondences required to solve for the pose from the $n$ available correspondences.
For robust pose estimation, this minimal solver is used inside a \acrfull{ransac} loop.
The \acrshort{pnp} problem for a \rs camera is explored by \cite{albl2015r6p}.
They show that the absolute pose problem (\acrshort{pnp}) can be solved using six 2D-to-3D correspondences.
They first use the traditional \acrfull{p3p} algorithm \citep{haralick1991analysis} to get an initial orientation estimate and then use the linearized motion model to account for \acrlong{rs}.
To solve for the motion, they use the Gr{\"o}bner basis method.
The Gr{\"o}bner basis technique is useful to solve a polynomial system of equations which naturally arise in the \acrshort{pnp} problem \citep{cox2006using}.
When the vertical direction is available, \eg via \acrshort{imu} in a smartphone, \cite{albl2016rolling} show that the \acrshort{pnp} problem can be solved using only five 2D-to-3D correspondences.
Similar to the previous approach, they use a Gr{\"o}bner basis method and the linear motion model, but do not require an initial orientation estimate while being more efficient.
\cite{saurer2015minimal} developed another minimal solver based on the Gr{\"o}bner basis.
They argue that when \rs cameras are used in cars, the dominant motion is translation, and hence a translation only model with linear velocity is appropriate.
They used five 2D-3D correspondences to estimate the translation and the unknown constant rotation.

\myparagraph{Relative pose problem}
\cite{Dai2016rollingEpipolargeom} developed epipolar constraints for \rs cameras and showed that the constraints do not define a line but rather define epipolar curves.
Analogous to the linear 8-point algorithm \citep{longuet1981computer} for \acrshort{gs} camera, they derive a 44-point linear algorithm for \rs camera to solve for a 7$\times$7 essential matrix assuming uniform rotation and translation.
For a purely translating \rs camera, they also propose a 20-point linear algorithm to solve for a 5$\times$5 essential matrix.
Using 44 points in a \acrshort{ransac} loop is impractical, and \cite{lee2017inertial} alleviate this using \acrshort{imu} readings and propose a linear solver using 11 points for a camera undergoing constant translational and rotational velocities.
When angular velocity dominates the linear velocity, only the rotational model is useful and \cite{lee2019gyroscope} develop a Gr{\"o}bner basis method to estimate an essential matrix using five correspondences, where the instantaneous rotation is obtained from the \acrshort{imu}.
\cite{zhuang2017rolling} estimate a differential relative pose for constant translational and rotational acceleration for \rs cameras.
Using optical flow to establish correspondences, they can estimate the relative pose using 9 correspondences.
For constant translational and rotational velocities, the same approach can be used to estimate the relative pose using only 8 correspondences.

\myparagraph{Multi-view stereo structure-from-motion}
\cite{saurer2013rolling} adapted the plane sweeping stereo algorithm \citep{yang2003multi, gallup2007real} for \rs and showed that the combination of RS artifacts and lens distortion leads to a biased 3D reconstruction if global shutter is na{\"i}vely assumed.
\acrshort{sfm} traditionally assumes \acrshort{gs} and was thus shown to be unreliable when used with images with large RS distortion \citep{liu2011subspace, hedborg2011structure}.
To resolve this, \cite{hedborg2012rolling} propose a \rs aware \acrfull{ba} method to account for the \rs effect.
This effort improved the \acrshort{sfm} pipeline, which had previously not worked well for \rs images with significant motion \citep{liu2011subspace}.
Later, \cite{saurer2016sparse} developed a full-fledged \acrshort{sfm} pipeline in which they assumed constant intra-frame rotational and translational velocity (similar to \cite{albl2015r6p}) during \acrshort{ba} and enforced temporal smoothness across frames using \acrfull{gps} readings.
\cite{albl2016degeneracies} also show that \rs cameras exhibit more degenerate configurations in \acrshort{sfm} as compared to the traditional pinhole \acrshort{gs} model.
In such cases, \acrshort{ba} prefers a trivial case where the scene is estimated to be planar.
\cite{ito2017self} also discovered such degeneracies while investigating self-calibration for a purely rotating \rs camera when the aspect ratio of the focal length and the skew parameter are unknown.
\myparagraph{\glsentrytext{rs} camera as a sensor}
Instead of serving as a source of error, the following methods treat the \rs effect as a source of information.
I follow this spirit in my \hf tracking approach described in ~\cite{bapat2016towards} and ~\cite{bapat_rolling_radial_tracking}.
\cite{ait2006exploiting} use a \rs camera to create a velocity sensor.
For an object with known geometry and a stationary \rs camera imaging it, they estimate the uniform translation and angular velocity of the moving object using sparse 2D-3D correspondences.
They solve for the motion by modeling the distortion in the image induced by the combination of object motion and the \rs effect.
\cite{magerand2012global} used constrained global optimization to estimate uniform motion of a known object captured using a \rs camera.
Similar to the previous work, they treated the \rs camera as a highly sensitive motion sensor and given the known object geometry and 2D-3D point correspondences, they estimate the absolute pose of the camera with respect to the object.
\cite{ait2007kinematics} extend the above point-based methods to use line correspondences.
These methods assume that the object geometry is known a-priori.
To alleviate this, \cite{ait2009structure} study the problem of 3D point triangulation and velocity estimation of a moving 3D object with unknown geometry using a \rs stereo rig.
Since all the points in the image are captured at different times, they assume a constant velocity of the 3D object and solve a non-linear optimization problem to estimate 3D locations as well as object velocity by minimizing the reprojection error.

\subsection{\glsentrytext{rs} calibration}
Rolling shutter exposes consecutive rows with a delay and hence the \rs camera needs to undergo time-delay calibration.
\cite{geyergeometric} propose a method to estimate the rate of rows exposed per second $n_r$ and the frame delay $t_f$.
If $n_r = \frac{H}{\mathrm{fps}}$ then by definition $t_f = 0$.
They estimate $n_r$ by flashing light with known frequency using a pulse generator onto a \rs camera without a lens.
They process the captured pattern of light and estimate $n_r$ using the Fourier analysis.
\cite{oth2013rolling} use just a video and forgo the use of special hardware but achieve higher calibration accuracy.
They use a weak motion prior with a continuous time trajectory model to estimate the line-delay $t_r$.
They parameterize the pose using a fourth order B-spline and iteratively update their model parameters while minimizing reprojection errors.
I use this method for \rs calibration.

\section{Tracking}
\label{sec:related_work_tracking}
High-frequency throughput has long been a main research thrust in the three primary technologies, namely tracking, rendering, and display,  of \acrshort{arvr} \citep{azuma1997survey,zhou2008trends}.
In this section, I will provide an overview of recent research efforts that have focused on addressing the tracking problem.
On a high level, methods addressing tracking can be divided into three categories: \begin{enumerate*} \item sensor-based, \item a hybrid of sensor- and visual-based, \item and purely visual-based.\end{enumerate*}

\subsection{Sensor-based tracking}
High-frequency sensor-based tracking systems have been deployed with some degree of success for many decades; see \cite{rolland2001survey} for a good review of early methods on this front.
More recently, \cite{lavalle2014head} have proposed to use only inertial sensors -- such as accelerometers, magnetometers, and gyroscopes -- combined with a predictive system for head pose tracking by assuming constant angular velocity or constant acceleration.
Intriguingly, they also report limited success in positional tracking, in addition to rotational tracking, using only the inertial sensors.
However, the inherent limitation of drift accumulation and the explosion due to integration of velocities to calculate the orientation renders the inertial sensors ineffective for full 6-\acrshort{dof} tracking.
As such, most recent work has focused on using sensors and vision systems jointly to perform tracking.

\subsection{Hybrid tracking}
One approach toward hybrid sensor/vision tracking is to use inertial sensors as the primary tracking component and augment them with vision systems to mitigate drift \citep{leutenegger2015keyframe}.
\cite{klein2004tightly} used predictions from \acrshort{imu} sensors to account for motion blur and built a parametric edge detector on video input to prevent drift.
\cite{persa2006sensor} used \acrshortpl{imu} and Kalman filtering for tracking, and to correct for drift, the author used \acrshort{gps} in outdoor environments and fiducial markers in indoor scenes.
\cite{mourikis2007multi} use the \acrshort{ekf} technique to incorporate geometric constraints for a single keypoint across multiple observations (images) whenever a new image is available, and in the meantime rely on the \acrshort{imu} for tracking.
\cite{li2013real} improve upon this \acrshort{ekf}-based technique by modeling a \rs camera by assuming a uniform translational and angular velocities.
All of these systems use the \acrshort{imu} as the workhorse, and camera-based computer vision technique is the secondary tool, trying to augment and correct the \hf \acrshort{imu} readings.

\subsection{Visual-based tracking}
\label{sec:related_work:tracking}
Visual trackers, which consider cameras as primary sensors, can be broadly classified into either feature-based or direct approaches.
Whereas the former analyze a set of salient image keypoints, the latter rely on  global image registrations.
The following discusses research for tracking in general, as well as \hf tracking in particular for feature and direct methods.
\subsubsection{Feature-based systems}
The early successful trackers like the \acrshort{klt} tracker \citep{tomasi1991tracking} and Shi-Tomasi tracker \citep{shi1993good} established criteria for quality of `ideal' feature points which results in robust tracking under a brightness constancy assumption among consecutive images.
With the innovations of more descriptive features like \acrshort{sift} \citep{lowe1999object} and \acrshort{surf} \citep{bay2006surf}, and efficient descriptors like \acrshort{orb} \citep{rublee2011orb}, \acrshort{fast} \citep{rosten2006machine}, and \acrshort{brisk} \citep{leutenegger2011brisk}, tracking with sparse features in real-time became viable.
Many feature-based tracking systems are subsystems of the larger  problem of \acrfull{slam} and here, I will highlight just the tracking subsystem.
A typical procedure used by the tracking subsystem in a \acrshort{slam} framework is as follows: For each new frame, \begin{enumerate*} \item extract features or keypoints, \item associate the keypoints (or a subset of) with the keypoints from the previous frame (or keyframe), and \item estimate the camera motion by minimizing reprojection error\end{enumerate*}.
The extracted features and \textit{keyframes} are arranged in a bipartite graph, where the edges indicate that a feature projects into a keyframe.
\cite{klein2007parallel} developed such a system called \acrfull{ptam} which used separate threads for tracking and mapping to solve the \acrshort{slam} problem.
For each new frame they predicted a pose prior using a motion model.
Using the prediction, they track the features using a coarse-to-fine approach by first estimating a rough pose and then refining it using high-resolution features in an image pyramid scheme.
\cite{ventura2014global} used a client-server model, where the server has a 3D model of the environment constructed offline.
A mobile phone acts as the client and runs a \acrshort{slam} system in local frame of reference, using the server for global registration and bundle adjustment.
\cite{forster2014svo} demonstrated a `semi-direct' approach running at 300 \acrshort{fps} on a consumer laptop.
This method uses photometric error between projections of 3D points in consecutive frames for motion estimation and employs \acrshort{fast} \citep{rosten2010faster} features for the mapping stage.

Other feature-based hybrid tracking systems use strategically placed fiducial markers for tracking distinct points in the environment \citep{zhang2002visual, naimark2005encoded}.
Typically, \acrfullpl{led}, beacons, or unique texture markers are used in this framework.
The Hi-Ball system \citep{welch2001high} is one such construction, in which blinking \acrshortpl{led} are placed on the ceiling of a capture environment.
A cluster of infrared cameras (the ``Hi-Ball'') observes the blinking pattern of \acrshortpl{led}, and the system uses strong triangulation constraints combined with motion prediction to provide highly accurate 6-\acrshort{dof} pose.
The major drawback of using such fiducial markers is that it becomes cumbersome as well as costly to place them throughout the environment, and, moreover, the tracking system is limited to the confines of the area in which the markers have been installed.
I use the Hi-Ball system to obtain ground truth for the real-world experiments and to validate the accuracy of my approach in Chapters ~\cite{bapat2016towards} and ~\cite{bapat_rolling_radial_tracking}.
Concurrent to this thesis, \cite{blate2019implementation} developed a low-latency tracker using a stereo-pair of duo-lateral photodiodes and custom hardware.
The \acrshort{hmd} is equipped with \acrshort{ir} \acrshortpl{led} which are observed by the externally stationed stereo-pair.
The duo-lateral photodiodes provide independent measurements in X and Y direction in the imaging plane, providing accurate positions of three \acrshortpl{led}, which enables pose estimation at 50kHz and 28$\mu$s.

Feature-based methods fundamentally limit themselves in utilizing image information which conforms to the definition of the feature.
In contrast, dense methods utilize the complete image for estimating the relative pose, but this comes at a much larger computational complexity, as described next.
\subsubsection{Direct/dense tracking}
Direct \acrshort{slam} approaches have also succeeded for both visual tracking and scene reconstruction in recent years.
\cite{newcombe2011dtam} used dense image registration to a 3D model to localize the camera, and used the registered image to update the 3D model in a loop.
The LSD--\acrshort{slam} algorithm of \cite{engel14eccv} and its derivatives \citep{caruso2015_omni_lsdslam, engel2015_stereo_lsdslam} use direct, semi-dense image alignment to reconstruct 3D models at nearly the rate of frame input.
For tracking, they minimize dense photometric error between the keyframe and the current frame.
To achieve this, they keep a rolling buffer of depth per keyframe which they refine using the adjacent frames.
In addition to the depth, they use error propagation to keep track of variance in depth estimates, which they utilize to normalize the dense photometric error.
This helps in giving importance to highly accurate 3D points over others.
\cite{schoeps14ismar} further introduced a direct approach on a mobile phone using semi-dense depth maps for a mesh-based geometry representation.
Their method is hybrid in that they find the ground plane using data from the built-in accelerometer on the mobile device.

Overall, direct \acrshort{slam} methods have been demonstrated to work on a large scale with relatively low computational requirements, and these systems have been leveraged for pose estimation in \acrshort{ar} systems.
However, the key limiting factor preventing these visual tracking methods from general \acrshort{arvr} use is that they require full camera video frame inputs, which is a substantial bottleneck for the overall system latency.
Effort by \cite{dahmouche2008high} towards this end was to predict and capture only the \acrfull{roi} around corner points/blobs as opposed to capturing complete image and then finding the edges/features. 
They predicted the \acrshortpl{roi} based on previous data under a constant velocity assumption to simultaneously track pose and velocity at 333Hz using a high speed camera. 

Recent direct methods incorporate \rs camera model in their motion estimatiion framework.
\cite{kim2016direct} introduced a monocular \acrshort{slam} system specifically for \rs cameras by adapting LSD--\acrshort{slam}~\cite{engel2014lsd}.
They parametrized the intra-frame motion via a k-control point B-spline.
In contrast to our method, they assumed no radial distortion and estimate single pose for the entire frame.
More recently, they incorporated radial distortion in \citep{kim2017rrd} by using generalized epipolar curves instead of epipolar lines.
Unlike \acrshort{slam} systems, my camera tracking system does not build an environmental map. 
Instead it relies on scan-line stereo depth variations across time to estimate 6 \acrshort{dof} motion, see ~\cite{bapat2016towards} for details.

In the papers ~\cite{bapat2016towards} and ~\cite{bapat_rolling_radial_tracking}, I have introduced a tracking system (without mapping) which essentially turns the rolling shutter camera into a \textit{computational sensor}.
I have demonstrated the effectiveness of \rs as a \hf measurement device.
This is possible because I explicitly estimate intra-frame motion.
Furthermore, I showed that treating the radial distortion as a virtue also benefits the tracker, and hence methods like \citep{kim2016direct, kerl2015dense, patron2015spline} which parameterize the intra-frame motion using splines can benefit from my tracking method.

\section{Edge-aware optimization}
\label{sec:related_work:edge_aware}
I will review the prior work related to edge-based filtering and optimization, namely: \begin{enumerate*} \item bilateral filters and their variants, \item optimizations leveraging superpixels, \item machine learning for edge-aware filtering,\item the domain transform and its filtering applications,\item and bilateral solvers.\end{enumerate*}
In the following, I will refer to an algorithm as a \textit{filtering technique} when a filter is applied on the input image to produce an output image.
On the other hand, I will refer to an algorithm as a \textit{solver} when it uses one or more input images and optimizes towards a goal defined by a cost or a loss function.
 
\subsection{Bilateral filters}
The bilateral filter was introduced by \cite{tomasi1998bilateral} and is one of the initial edge-aware blurring techniques. The major bottleneck for bilateral filtering is that it is costly to compute, especially for large blur windows.
For $N$ pixels in an image, filter radius $r$ in each dimension, and $d$ dimensions, its complexity is $\mathcal{O}(Nr^d)$ \citep{pham2005separable}.
Since its invention, there have been multiple approaches proposed to speed up the bilateral filter \citep{weiss2006fast, adams2010fast, chen2007real, yang2015constant, elad2002origin, durand2002fast, pham2005separable, paris2006fast}.
\cite{durand2002fast} approximate the bilateral filter by a piece-wise linear function.
\cite{pham2005separable} proposed to use two 1-D bilateral kernels, reducing the complexity to $\mathcal{O}(Nrd)$.
\cite{paris2006fast} treat the image as a 5-D function of color and pixel space and then apply 1-D blur kernels in this high-dimensional space, leading to complexity of $\mathcal{O}(N + N/r^2\times \prod_{i=3}^{d}(D_i/r))$, where the $N/r^2$ term is due to the 2-D pixel space and the remainder is the contribution of dimensions ${i =3 \dotsc d}$, each with size $D_i$, \eg $D = 255$ for an 8-bit grayscale image \citep{paris2009bilateral}.
Note that this is still exponential in the dimensionality \citep{adams2010fast}.
These approximations to the bilateral filter decouple the 2-D adaptive bilateral kernel into a 1-D kernel, reducing the computational cost significantly.

When used as a post-processing step, the bilateral filter removes noise in homogeneous regions but is sensitive to artifacts such as salt and pepper noise \citep{zhang2008multiresolution}.
The method introduced in~\cite{bapat2018_dts} emphasizes edge-aware concepts in the same spirit as bilateral filters, and the formulation yields a generalized optimization framework that is efficient and accurate.
When using robust cost functions, my method remains resistant to outliers.

\subsection{Superpixels}
To combat issues of computational complexity during bilateral optimizations, several approaches leverage superpixels, \textit{i.e.}, they group pixels together based on appearance and location.
These approaches treat the superpixels as atomic unit which are far fewer in number in contrast to pixels, reducing the processing time.
Superpixel extraction algorithms like \acrfull{slic} \citep{achanta2012slic} are often used in optimization problems for two major reasons: 1) They reduce the number of variables in the optimization \citep{cadena2014semantic}, and 2) they adhere to color and (implicitly) object boundaries.
In one application, \cite{bodis2015superpixel} use sparse \acrshort{sfm} data, image gradients, and superpixels for surface reconstruction to ensure that the edges of triangles are aligned to the edges in the image.
\cite{lu2013patch} use \acrshort{slic} superpixels to enforce spatially consistent depths in a PatchMatch-based matching framework \citep{barnes2009patchmatch} to estimate stereo.
As superpixels may cover arbitrarily large regions with widely varying depths, unfortunately, using them leads to a loss in resolution, as an entire superpixel is assigned one estimate; \eg, in stereo, this leads to fronto-parallel depths.
Additionally, using superpixels inherently assumes perfect segmentation, and these assumptions often do not hold in practice.

\subsection{Machine learning for edge-awareness}
\cite{porikli2008constant} achieves independence to kernel sizes using a taylor series approximation, but is inaccurate for low color variances.
To remedy this, \cite{yang2010svm} use \acrfullpl{svm} to mimic a bilateral filter by using the exponential of spatial and color distances as feature vectors to represent each pixel.
Their approach supports varying intra-image color variance while remaining efficient.
Traditionally, \acrfullpl{crf} are used for enforcing pair-wise pixel smoothness via the Potts potential.
Alternatively, \cite{krahenbuhl2011efficient} proposed to use the permutohedral lattice data structure \citep{adams2010fast}, which is typically used in fast bilateral implementations, to accelerate inference in a fully connected \acrshort{crf} by using Gaussian distances in space and color. 

With the recent explosion of compute capacity and convolutional models in the vision community, there are also deep-learning methods that attempt to achieve edge-aware filtering.
\cite{chen2016deeplab} presented DeepLab to perform semantic segmentation; there, they use the fully connected \acrshort{crf} from \cite{krahenbuhl2011efficient} on top of their \acrshort{cnn} to improve the localization of object boundaries.
\cite{xu2015deep} learn edge-aware operators from the data to mimic various traditional handcrafted filters like the bilateral, weighted median, and weighted least squares filters \citep{farbman2008edge}.
More recently, deep learning methods have been introduced to learn optimal bilateral weights \citep{gharbi2017deep, liu2017learning, wu2018fast} instead of using the traditional Gaussian color weights.
However, machine learning approaches require large amounts of training data specific to a task, plus significant compute power, while my approach works without any task-dependent training and runs efficiently on a single \acrfull{gpu}, see ~\cite{bapat2018_dts} for details.

\subsection{The domain transform}
\cite{gastal2011domain} introduced the domain transform, a novel and efficient method for edge-aware filtering that is akin to bilateral filters.
The domain transform is defined as a 1-D isometric transformation of a multi-valued 1-D function such that the distances in the range and domain are preserved.
When applied to a 1-D image with multiple color channels, the transformation maps the distances in color and pixel space into a 1-D distance in the transformed space.
When the \textit{scalar} distance is measured in the transformed space, it is equivalent to measuring the \textit{vector} distance in [R, G, B, X] space. 
This has the benefit of dimensionality reduction, leading to a fast edge-aware filtering technique which respects edges in color while blurring similar pixels. 
To apply the domain transform to a 2-D image, the authors apply two passes, one in the X direction and one in Y.
This has a complexity of $\mathcal{O}(Nd)$, and as a result scales well with dimensionality.
\cite{chen2016semantic} proposed to perform edge-aware semantic segmentation using deep learning and use a domain transform filter in their end-to-end training of their deep-learning framework.
They also alter the definition of what is considered as `edge' by learning an edge prediction network, and they then use the learned edge-map in the domain transform.
Their application of the domain transform is in the form of a \textit{filter}.

Applying the domain transform to an image results in a filtering effect, in contrast to my method which optimizes according to an objective function and hence is a solver.
I use the domain transform in my method in an iterative fashion within the optimization framework because it provides an efficient way to compute the local edge-aware mean.
The application of the domain transform as a filter similar to \cite{chen2016semantic} produces less accurate results than my framework.
See ~\cite{bapat2018_dts} for more details.

\subsection{Bilateral solvers}
Recently, \cite{barron2015fast} suggested to view a color image as a function of the 5-D space [Y,U,V,X,Y], which they call the `bilateral space', to estimate stereo for rendering defocus blur in a mobile phone.
They proposed to transform the stereo optimization problem by expressing the problem variables in the bilateral space and then optimizing in this new space.
I will refer to this method as \acrfull{bls}. 
\cite{barron2016fast}'s \acrfull{fbs}, on the other hand, solves a linear optimization problem in the bilateral space, which is different from \acrshort{bls}.
In this setting, they require a target map to enhance, as well as a confidence map for the target.
The linearization of the problem allows them to converge to the solution faster.
Both of these approaches quantize the 5-D space into a grid, where the grid size is governed by the blur kernel size.
This reduces the number of optimization variables and hence the complexity, leading to low runtimes.
\cite{mazumdar2017hardware} use a dense grayscale-space 3-D grid (instead of a 5-D color-space grid) and the Heavy-Ball algorithm to make \acrshort{fbs} faster, but at the cost of accuracy.
I compare my method with the above solvers in ~\cite{bapat2018_dts}, and show that my method is more efficient than existing techniques.

\chapter{Limitations and future directions}
\label{chap:future_work}
Through extensive experimentation, I have shown the effectiveness of the \hf tracking methods in ~\cite{bapat2016towards} and  ~\cite{bapat_rolling_radial_tracking} and that of the efficient edge-aware optimization algorithm in ~\cite{bapat2018_dts}.
The work in this dissertation has limitations which can be improved upon in various ways for future research, some of which are outlined here.
Accompanying the promising research direction, I have also provided the biggest hindrances that I think might hamper progress.

\section{High-frequency tracking}
This thesis outlines a prototype method to address \hf tracking.
Although, this method is quite successful in tracking at high-frequencies, following are the limitations and directions of research we can build upon it in the future.
I will first address the limitations of the proposed approaches and then describe the future research directions one can explore specific to my system.
The \hf tracker proposed in this thesis should be realized using \rs cameras providing row-level synchronization.
Towards, that end I have provided a sketch design assuming such a \rs synchronization is available, and I discuss promising camera candidates and hardware useful for building such a system.
If such a \rs synchronization is not possible, another direction is to explore more complex lenses on a single \rs camera.
Using a multi-lens on a \rs camera effectively creates virtual cameras which are implicitly synchronized, and I will describe this envisaged tracker.

There are general issues in the space of \hf tracking, which still remain unsolved.
An ideal tracking system functions well in all environments, has low-latency and \hf, is robust and is ergonomically suitable for daily wear.
\cite{welch2002motion} argue that {no single tracking technique is likely to emerge to solve the problem of every technology and application}.
To fulfill the above stated qualities of the tracking system, we need a better understanding of how much latency and \hf is required for comfortable daily wear.
Additionally, we need to design robust systems which work outdoors as well indoors, and are robust to changing lighting conditions and dynamic elements in the scenes.
Moving forward, we need to also recognize that multiple users are going to use \acrshort{arvr} in the same space.
Following the old adage `\textit{The whole is greater than the sum of its parts}', an interesting research direction is how to complement and help multiple trackers, where instead of interfering, the trackers help each other.
I describe all of these themes in more detail in the following.

\subsection{Row-matching}
The \hf tracking algorithms depend upon estimation of pixel shifts using the binary descriptors.
Since the binary row-descriptor uses the image gradient, it is implicitly assumed that the scene has sufficient texture.
Hence, the tracker will fail in a completely textureless scene.
If a user is situated in a completely textureless room,~\eg a completely white room, an interesting question to ask is `Is tracking even necessary in such a case since the user also cannot localize visually?'.
While the user has no discriminating visual signal, the vestibular system which provides humans with sense of position and orientation (proprioception), other signals like the sense of position due to sound reverberations and the self-awareness about the movement of the head/neck before we tell the brain to move, all should indicate that the head has (will) move.
A simple example of such human sense is evident in the children's game `Blind man's buff' where a blind-folded person is rotated to confound their sense of direction and is asked to catch the other players.
In that game, although we lose the sense of absolute position and orientation, we do retain relative sense of position and orientation.
This intuitively warrants head-pose tracking for \acrshort{arvr}, although the precise requirements in terms of accuracy need to be studied by conducting user studies.
\subsection{Frame overlap and occlusion}
Additionally, the success of row matching assumes that the same 3D point is observed in the two row-images.
This assumption is violated when the 3D point is occluded by a foreground object.
Occlusion depends upon the combination of the camera motion and scene geometry and is difficult to avoid.
There are no certifiable guarantees that the matching is correct and occlusion has not happened.
An incorrect match due to occlusion cannot be disambiguated by reconstructing the surrounds and then ray-casting since we need to know the motion.
Additionally, the motion compensation technique based on homography-based warps (see Our Approach section in ~\cite{bapat2016towards} and ~\cite{bapat_rolling_radial_tracking}) rely on the fact that the current row is previously observed in the past image data.
The defense used in the proposed methods against such incorrect matches is to modulate the confidence scores according to the smoothness of the estimated shift and the as well as to used robust filters.
Additionally, the hope is that the redundant cameras observe the motion, even if some of the cameras in the cluster face strong occlusion or textureless regions.
For robustness, one direction to explore is to use a \acrshort{ransac} framework to estimate the minimal solution by randomly selecting the constraints.
Furthermore, a bunch of scan-lines can be used instead of a single row for increases robustness at the cost of some latency.
\subsection{Dense sampling of linear segments}
The virtual \rs cameras induced by the radial distortion correspond to a linear line-segment in the undistorted space (refer ~\cite{bapat_rolling_radial_tracking}).
Using the constraints induced by the virtual cameras, a over-determined linear system is solved to estimate the head motion.
A natural extension is to the sample the line-segments densely, effectively creating a segment per pixel of the row.
The use of such approach is akin to the dense variants of \acrshort{slam}. 
\subsection{Drift correction}
The proposed approaches drift due to the absence of any drift correction mechanism.
The \acrshort{slam} systems use a keyframe-based framework to reduce such drift where the relative poses between keyframes are optimized \citep{kummerle2011g2o}.
The presented methods will benefit by graph-based optimization scheme to reduce the accrued drift.
It needs to be investigated how often such drift correction is required and what is a `good' keyframe in the presence of \rs.

\subsection{Exposure length}
The exposure of the row effectively blurs the intra-row motion for the $t_e$ time for which the row is exposed.
This is the same as running a low-pass filter of width $t_e$ on the observations and in turn the observed motion.
For a 120 Hz camera, the maximum exposure length is 8.33 ms which is too long for the latency requirements for \acrshort{arvr}.
The requirements for tracker latency is given by the motion-to-pose latency which is much shorter than the \acrfull{mpl}.
The \acrshort{mpl} is in the range of a few milliseconds \citep{zheng2014minimizing}.
Hence, the \rs camera should be exposed for much smaller exposure times to reduce the use of stale pose information.
This implies that the tracker functions better in well-lit scenes as it can achieve smaller motion-to-pose latency.
This motivates further study into the effect of exposure length on the tracker latency.

\subsection{Hardware implementation: A sketch}
\begin{figure}
    \centering
    \includegraphics[width=0.6\textwidth]{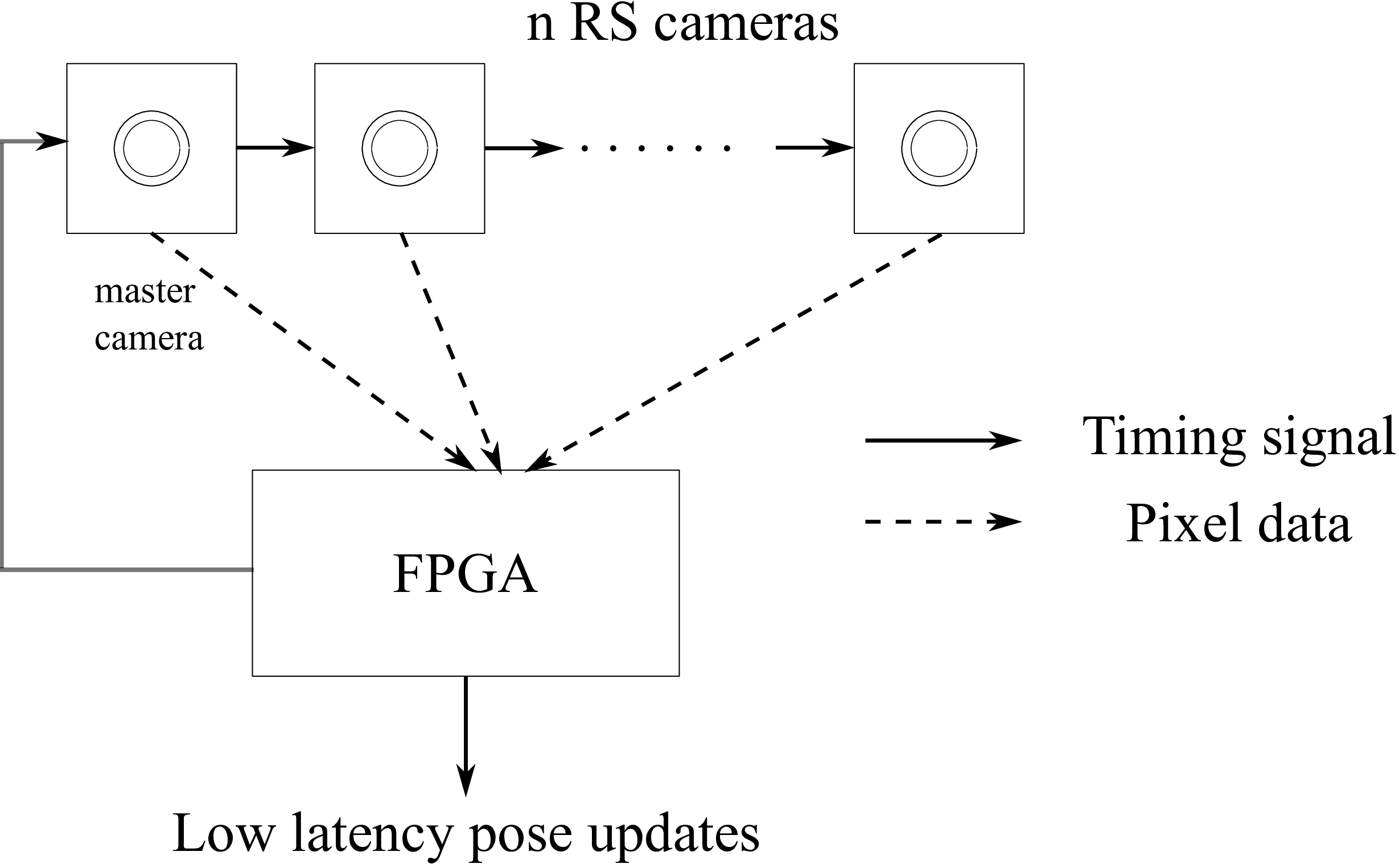}
\caption{\label{fig:hardware_sketch} The \acrshort{fpga} sends out control signal to master camera, which in turn drives the exposure in the $n$ \acrshort{rs} of the camera cluster. The cameras in turn send a row of the image as soon as the row is readout from the sensor which is processed by the \acrshort{fpga} to give \hf and low-latency 6 \acrshort{dof} pose updates.}
\end{figure}

The next step to build upon this thesis is to implement the system in hardware.
The following provides a sketch to achieve this objective.
First we need a \acrshort{rs} camera which provides hardware row-level sync as well as access to row-level pixel data with minimal latency.
Suggested cameras are OV9740, OV6946, OV6948 and OV6922 all by OmniVision, with the most promising being OV6930 and the Iris 15 by Teledyne Photometrics.
Additionally, we need a \acrfull{fpga} or an \acrfull{asic} board capable of receiving multiple video streams from multiple cameras and logiADAK Zynq-7000 \acrshort{soc} by Xilinx seems to be promising for this purpose.
With this hardware in mind, an implementation as shown in Fig. \ref{fig:hardware_sketch} is possible.
The $n$ cameras in the cluster are arranged in a master-slave fashion where the master camera is triggered by the \acrshort{fpga}.
All the cameras synchronously capture a row and send out the row-pixel data as soon as it is read from the sensor to the \acrshort{fpga} for further processing.
The \acrshort{fpga} deploys the \hf tracking algorithm to produce pose updates.

\subsection{Tracking using multi-lens arrays}
\begin{figure}
    \centering
    \includegraphics[width=0.6\textwidth]{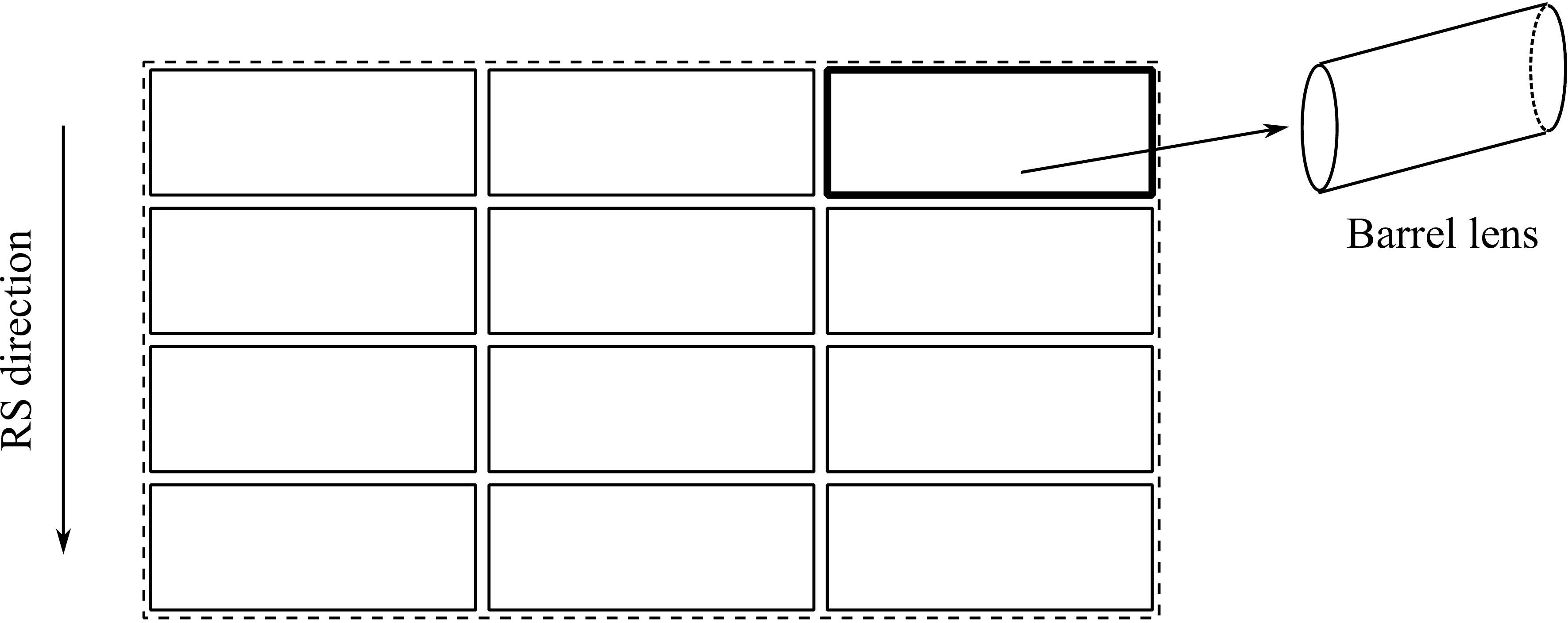}
\caption{\label{fig:multilens_tracker} The multiple barrel lenses focus large parts of the scene on a subset of the rows increasing the descriptiveness of the rows as well as provides row-level sync for free across the lenses.}
\end{figure}

\cite{bishop1984self} used a barrel lens to project the entire scene onto a line-camera in their tracker.
Motivated by this use of a barrel lens, the work described in~\cite{bapat_rolling_radial_tracking} and micro-lens arrays typical of light-field cameras, an interesting direction to pursue is to use a more complex lens on a single \acrshort{rs} camera.
As noted before, one of the main challenges to a multi-camera tracker is obtaining row-level sync.
This constraint can be overcome by using a multi-lens array which focuses large overlapping chunks of the scene onto individual rows of the camera using barrel distortion.
One such design is shown in Fig. \ref{fig:multilens_tracker}.
Each smaller rectangle is a barrel lens with large \acrshort{fov} along the vertical direction, but small \acrshort{fov} along horizontal direction.
This in effect is an array of virtual cameras which have synchronous row-exposures. Additionally, the barrel distortion increases the \acrfull{snr} for the matching of the row-descriptors.
Hence, a single camera can provide synchronous observations, albeit with small baseline.
The error in depth is inversely proportional to the baseline \citep{gallup2008variable}, and the small baseline in the observations can be a cause of concern.
Moreover, this method can be aided by using an \acrshort{ir} projector system similar to Microsoft's Kinect so that the tracker works in texture-less environments as well.

\subsection{Tracker latency requirements and measurement}
There has been much research attempting to quantify typical daily human motion \citep{bussone2005linear,list1983nonlinear}, as well as extreme motion, \eg in an American football game \citep{rowson2009linear}, the latency requirements for tracking such range of motions \citep{bailey2004latency,jerald2010scene,jerald2009relating} and measurement accuracy of trackers \citep{allen2007hardware, vorozcovs2006hedgehog, wiersma2013spatial, nafis2006method, pustka2010determining}.
Although it is clear that higher tracking frequency and lower latency is required \citep{pausch1992literature}, there is no definitive answer as to how much frequency is enough.
This in part is due to the fact that we need \hf trackers to measure physiological response at such low latency and high motion in the first place.
The tracker by \cite{blate2019implementation}, although limited in tracking volume and free movement, might prove to be useful for obtaining such measurements.
There is a need to measure how much \acrshort{mpl} is required, and in particular how much latency is acceptable for tracking systems.
One measure to quantify this is the \acrfull{jnd} for visual acuity in static as well as dynamic environments.

\subsection{Generalizability and robustness}
An ideal tracking system has at least the following qualities: \begin{enumerate*}
\item works inside closed spaces as well as outdoors, \ie is generalizable,
\item low-latency and \hf,
\item robust to sudden changes and to dynamic objects in the environment, and
\item ergonomics: is low-power, tether-less and has form-factor of eye-glasses.
\end{enumerate*}
Achieving all of these criteria in a single tracker is often treated as a dream \citep{bishop1995tracking, welch2002motion}, and rather we are always in a multi-dimensional trade-off space.
Multiple issues affect the generalizability and robustness of the tracker, \eg high-dynamic range of the scene outdoors \vs indoors, depth variability and complexity of the scene, range of possible human motion and dynamic objects in the environment, to name a few.
Improving on these issues is a promising direction of research.

\subsection{Social tracking}
\textit{Social} tracking is the natural extension to \hf tracker, where multiple users of \acrshort{arvr} are situated in a common physical space and share data to improve tracking.
When multiple users share a space simultaneously, robustness against dynamic objects and occlusions becomes crucial.
Not only do we need to resolve interference among trackers, \eg HTC Vive lighthouse tracking interferes with the Hi-Ball tracker, but we need to design trackers which can collaborate, resulting in better overall robustness.
One way to achieve this is to jointly share and update the 3D model of the shared space for a \acrshort{slam} system \citep{kim2010multiple, andersson2008c}.
Such a sharing of maps is also beneficial even if the users are not co-located at the same time.
As a new user explores the 3D space, they can reuse previously captured data by some other user.
A major challenge in this envisaged use case is to fuse observations from different sensors and modalities which are captures asynchronously.
This asynchronicity can be helpful to get even higher tracking frequencies via the use of the \acrfull{scaat} algorithm \citep{welch1997scaat}, which allows the individual sensors to capture at lower rates with system updates at higher rates.
Additionally, the trackers should be able to model any errors caused by the presence of the other users and interference caused by their tracking devices.

One example of tracking systems helping each other is the case of co-located users at the same time, and if the tracking systems have active components, \eg lasers or \acrshort{ir} \acrshortpl{led}, these can be used as additional external beacons to aid in tracking.
A complex of social tracking is where most people are using an \acrshort{arvr} device.
Even if the individual trackers might be insufficient to track by themselves, all the trackers present in the shared space together enable tracking for all.
Communication overhead introduced by the data sharing required for enabling the above is another separate challenge.

\section{Edge aware optimization}
The edge-aware optimization problem described ~\cite{bapat2018_dts} utilizes the reference (guide map) which is defined on a regular pixel grid, to define edges in N-dimensional space.
Using the edges, the solver minimizes a cost function, where the input target is an initial solution.
The work on efficient edge-aware optimization can be advanced further by generalizing each of these characteristics of the problem.
The method can be extended to  \begin{enumerate*} \item different domains on which the reference (guide map) is defined, \item what is being optimized/estimated, \item how to characterize or define an `edge', and \item how to apply edge-aware optimization to other domains\end{enumerate*}.    
Each of these characteristics of the problem is considered next.

\subsection{Irregular domains}
The domain for 2-D images is a regular pixel grid, limiting the reference to be dense and regular.
Often, the reference is irregular, sparse or is a surface/manifold in a higher dimensional space \citep{GastalOliveira2012AdaptiveManifolds}.
3D modeling and reconstruction is such an example of a surface, where most of the space is empty.
One can extend the work presented in ~\cite{bapat2018_dts} by adding additional \textit{sparse} depth (Z) dimension, effectively creating a hybrid domain with sparse and dense dimensions.
This can be further extended to general graphs, where neighborhood of vertices (pixels) are defined by adjacency lists.

Additionally, an irregular grid is useful for real-time applications like autonomous driving where we need to trade-off speed \vs time.
In real-time scenarios, we can preferentially process different regions of an image at different resolutions to save time, \eg the road is processed at higher resolution and sky at lower resolution, by modeling the image as an irregular grid.
Another example of an irregular grid is the equi-rectangular image used to represent $360^{\degree}$ images.
\cite{tateno2018distortion} highlight the need to model the unequal projection of the pixels in equi-rectangular image.

The above examples have irregular grids which are static across time.
Often, it is useful to apply a variable blur across time, \eg in a video, we might want to more aggressively blur static regions than dynamic regions, resulting into a dynamic irregular grid.
\subsection{Alternating optimization}
The initial target is the result of another optimization function, \eg depth in stereo problem.
One interesting direction of research is when the reference (guide map) itself is the product of another optimization.
For such a scenario, a solution is to alternate between four optimization problems by interchanging the role of target and reference: \begin{enumerate*} \item estimate target, \item edge-aware refine target with constant reference, \item estimate reference, and \item edge-aware refine reference with constant target.\end{enumerate*}
This alternating approach simultaneously estimates the reference and the target which are aligned along their edges.
This is useful in simultaneous refinement of geometry and texture \citep{bodis2015superpixel}.

\subsection{Semantics and \glsentrytext{cnn}}
Many machine learning tasks require pixel-level accurate predictions, \eg semantic segmentation, instance segmentation or masks, depth estimation and image synthesis  \citep{chen2016deeplab}. 
\acrshort{fbs} and \acrshort{dts} both can be embedded inside \acrshortpl{cnn} as a layer.
These layers can either serve as refinement on top of the prediction to make it edge-aware, or they can process deep features in an edge-aware sense to create better feature representations.
But, it is unclear whether \acrshortpl{cnn} need such edge-aware layers as refinement.
In some cases \acrshortpl{cnn} produce high-quality results which are aligned along edges \citep{he2017mask} and sometimes fail completely \citep{godard2017unsupervised}, which leaves another research direction to pursue.

Another interesting direction is to investigate what qualifies as an `edge'.
Typically, color edges are used, but depth or semantic disparities can provide stronger cues.
Simply utilizing semantic labels or depth as additional dimensions along with color-pixel 5D space is a simple extension. 
But, more interesting is to address whether the individual semantic labels constitute additional dimensions, or blurring across labels has any meaning?
For example, it does not make sense to combine labels `sky' and `trees', but it might be useful to combine `chair' and `stool' labels.
The confusion matrix of the algorithm which predicts the semantic labels might be useful to study how we can combine across labels.
This allows us to model complex interactions, where the dimensions of the bilateral space can be hybrid, \eg `foreground-depth', instead of two individual dimensions: `foreground' and `depth'.

\subsection{Data compression}
Telepresence \citep{wurmlin20043d} and streamable free view-point videos \citep{collet2015high} require good compression of color and depth or mesh data.
Since edge-aware optimization is quite accurate and efficient for depth super-resolution and colorization tasks, the natural extension is to use it in data compression.
One envisaged way to accomplish this is to sample a minimal number of points in the depthmap and create a sparse target, such that upon edge-aware processing we can recreate the original dense target with minimal artifacts.

\newpage
\bibliographystyle{apalike}
\bibliography{backmatter/domain_transform_references,backmatter/general_references}

\end{document}